\documentclass{article}

\usepackage{microtype}
\usepackage{graphicx}
\usepackage{subcaption}
\usepackage{booktabs} 

\usepackage{hyperref}
\usepackage{enumitem}

\usepackage[preprint]{icml2026}

\usepackage{amsmath}
\usepackage{amssymb}
\usepackage{mathtools}
\usepackage{amsthm}

\usepackage[capitalize,noabbrev]{cleveref}

\usepackage{float}
\usepackage{booktabs}
\usepackage{makecell}
\usepackage{graphicx}
\usepackage{xcolor}
\usepackage{booktabs}
\usepackage{multirow}
\usepackage{tabularx}
\usepackage{makecell}
\usepackage[table]{xcolor}
\usepackage{float}

\newcommand{\Rstruct}{\ensuremath{R_{\text{struct}}}}
\newcommand{\Rcont}{\ensuremath{R_{\text{cont}}}}
\newcommand{\Rlen}{\ensuremath{R_{\text{len}}}}
\newcommand{\Rout}{\ensuremath{R_{\text{out}}}}

\theoremstyle{plain}

\theoremstyle{definition}

\theoremstyle{remark}

\usepackage[textsize=tiny]{todonotes}

\icmltitlerunning{Social-R1: Towards Human-like Social Reasoning in LLMs}

\begin{document}

\twocolumn[
  \icmltitle{Social-R1: Towards Human-like Social Reasoning in LLMs}

  \begin{icmlauthorlist}

  \icmlauthor{Jincenzi Wu}{cuhk}
  \icmlauthor{Yuxuan Lei}{msra}
  \icmlauthor{Jianxun Lian}{msra}
  \icmlauthor{Yitian Huang}{msra}
  \icmlauthor{Lexin Zhou}{princeton}
  \icmlauthor{Haotian Li}{msra}
  \icmlauthor{Xing Xie}{msra}
  \icmlauthor{Helen Meng}{cuhk}

  \end{icmlauthorlist}

  \icmlaffiliation{cuhk}{The Chinese University of Hong Kong}
  \icmlaffiliation{msra}{Microsoft Research Asia}
  \icmlaffiliation{princeton}{Princeton University}

  \icmlcorrespondingauthor{Jianxun Lian}{jianxun.lian@microsoft.com}



  \vskip 0.3in
]

\printAffiliationsAndNotice{}  

\begin{abstract}
While large language models demonstrate remarkable capabilities across numerous domains, social intelligence—the capacity to perceive social cues, infer mental states, and generate appropriate responses—remains a critical challenge, particularly for enabling effective human-AI collaboration and developing AI that truly serves human needs.  Current models often rely on superficial patterns rather than genuine social reasoning.  We argue that cultivating human-like social intelligence requires training with challenging cases that resist shortcut solutions. To this end, we introduce \textbf{ToMBench-Hard}, an adversarial benchmark designed to provide hard training examples for social reasoning. Building on this, we propose \textbf{Social-R1}, a reinforcement learning framework that aligns model reasoning with human cognition through multi-dimensional rewards. Unlike outcome-based RL, Social-R1 supervises the entire reasoning process, enforcing structural alignment, logical integrity, and information density. Results show that our approach enables a 4B parameter model to surpass much larger counterparts and generalize robustly across eight diverse benchmarks. These findings demonstrate that challenging training cases with trajectory-level alignment offer a path toward efficient and reliable social intelligence.
\end{abstract}

\section{Introduction}

Recent advances in Reinforcement Learning from Verifiable Feedback (RLVF)~\cite{wen2025reinforcement} have significantly improved large language models (LLMs)' performance on formal reasoning tasks like mathematics and programming. However, genuine social intelligence—the capacity to perceive subtle cues, infer latent mental states, and navigate complex interpersonal dynamics—remains a substantial challenge. Unlike formal reasoning with deterministic execution paths, social reasoning is inherently polysemic and context-dependent, lacking the traceable verification that enables effective reward modeling in objective domains.

\begin{figure*}[t]
    \centering
    \includegraphics[width=\textwidth]{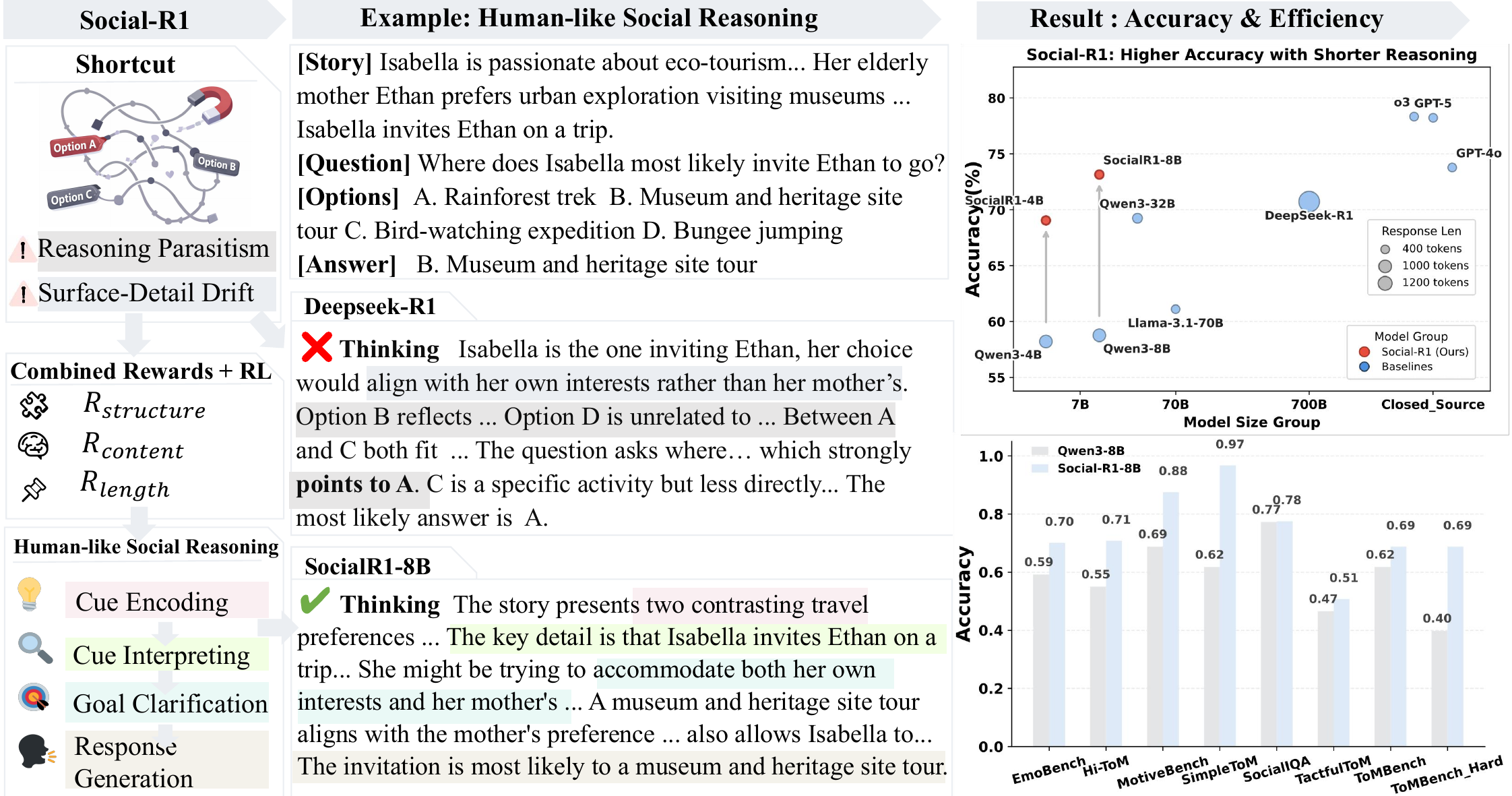}
    \caption{Social-R1 for Human-like and Efficient Social Reasoning. By integrating SIP-guided rewards into reinforcement learning, Social-R1 mitigates reasoning shortcuts and enforces structured human-like social inference, improving both accuracy and efficiency across model scales. Detailed cases are in Appendix \ref{app:0mainfigure}.}
    \label{fig:socialr1}
\end{figure*}

Despite strong performance on standard benchmarks, current LLMs often rely on shortcut learning rather than authentic social reasoning. Models frequently exhibit a failure mode we term \textit{Reasoning Parasitism}, characterized by \textit{Answer-driven Backfilling}—retroactively constructing justifications for predetermined answers rather than deriving inferences through narrative analysis, as illustrated by the example in Figure~\ref{fig:socialr1}. The inherent fragility of this parasitic performance becomes particularly evident in adversarial or out-of-distribution scenarios. As highlighted by \citet{shapira2024clever,pang2025large}, models that excel on standard benchmarks often suffer catastrophic failures when confronted with trivial narrative perturbations. It reveals that current approaches create only a facade of social intelligence while failing to develop robust reasoning capabilities. Our analysis further identifies a critical \textit{Interpretation Bottleneck}: while models can perceive surface-level social cues, they struggle to map these cues to latent mental states, leading to a ``logic reversal'' where final answer correctness exceeds the logical integrity of the reasoning process.

We argue that advancing social intelligence requires {aligning models' reasoning trajectories with the structured stages of human social inference}. Thus, we propose \emph{process-based trajectory alignment} that cultivates social intelligence as an internalized capability rather than a parasitic performance. Human social reasoning is characterized by high-density information distillation and recursive belief modeling—properties we aim to instill through structured supervision of the reasoning process, as illustrated in the {SocialR1-8B} example in Figure~\ref{fig:socialr1}.

In this paper, we introduce \textbf{Social-R1}, a reinforcement learning framework that encourages genuine social reasoning by aligning model trajectories with human cognitive principles: \emph{structured}, \emph{evidence-grounded}, and \emph{efficient} inference. Our approach has two key components. First, we construct \textbf{ToMBench-Hard}, an expert-curated adversarial benchmark designed to expose reasoning shortcuts through socially nuanced perceptual traps. Second, guided by \textbf{Social Information Processing (SIP)} theory \citep{salancik1978social} , we develop a multi-dimensional reward system that enforces stage-consistent reasoning progression ($R_{\text{struct}}$), content integrity ($R_{\text{content}}$), and inference efficiency ($R_{\text{len}}$). We demonstrate that this trajectory-level alignment enables more effective and parameter-efficient social intelligence.

We conduct comprehensive experiments across both in-domain and out-of-domain benchmarks related to social intelligence. Results demonstrate that our approach substantially improves model performance on genuine social reasoning tasks. Social-R1 enables smaller models to match or surpass the capabilities of models with significantly larger parameter counts, while showing strong generalization across eight diverse social reasoning evaluations (as illustrated in the right part of Figure~\ref{fig:socialr1}). These findings indicate that trajectory-level alignment presents a viable and efficient alternative to reliance on pure model scaling for achieving robust social intelligence.
The source code and dataset will be released upon paper acceptance.
In summary, our contributions are threefold:
\begin{itemize}[leftmargin=*]
    \item \textbf{ToMBench-Hard:} A rigorous benchmark with difficulty for social reasoning that exposes shortcut learning in LLMs and mandates genuine cognitive engagement.
    \item \textbf{Social-R1 Framework:} A reinforcement learning approach with multi-dimensional rewards that align LLM reasoning trajectories with human social cognition.
    \item \textbf{Performance Superiority:} Demonstration that our method enables small models to achieve large-model performance, proving trajectory quality surpasses parameter scaling for social intelligence.
\end{itemize}

\section{Related Work}

\subsection{Theory-of-Mind Analysis in LLMs}
\label{related:TOM}

Recent evaluations demonstrate LLMs' growing proficiency on Theory-of-Mind tasks, with models like GPT-4 achieving near-human performance on false-belief tasks and various social cognition assessments \cite{kosinski2024evaluating,street2024llms,strachan2024testing}. However, broader audits reveal significant limitations in genuine social reasoning capabilities. Studies highlight inconsistent performance across different evaluation frameworks \cite{gandhi2023understanding}, systematic failure modes in comprehensive capability assessments \cite{chen2024tombench}, and particular weaknesses in psychologically complex scenarios compared to physical-world contexts \cite{xu2024opentom,gu2024simpletom,zhou2023far}. Dynamic evaluation settings further expose models' difficulties with evolving mental states over time \cite{xiao2025towards}, while higher-order reasoning tasks reveal sharp performance drops beyond first-level inference \cite{he2023hi}.
Current improvement attempts remain limited in scope. Prompt-based interventions \cite{gu2024simpletom} offer task-specific benefits but lack generalizability, while reinforcement learning approaches \cite{lu2025tom} show promise but risk overfitting to narrow task patterns without broader training diversity.

From this literature, we identify three key limitations motivating our work: (i) over-reliance on outcome-based evaluation rather than reasoning trajectory analysis; (ii) susceptibility to shallow statistical cues instead of recursive mental-state inference; and (iii) limited generalization across social reasoning benchmarks and related tasks \cite{sabour2024emobench, yong2025motivebench}. These gaps necessitate harder adversarial evaluations, trajectory-level supervision, and training regimes designed for transferable social intelligence.

\subsection{RL for LLM General Reasoning}
Reinforcement learning has emerged as a pivotal technique for enhancing LLM reasoning capabilities, with RLHF establishing the foundation for alignment \cite{ouyang2022training}. Recent large-scale implementations demonstrate that outcome-based rewards can significantly strengthen structured reasoning in mathematical and coding domains \cite{jaech2024openai,guo2025deepseek,team2025kimi,yang2025qwen3}. To address sparse reward challenges, process supervision methods provide denser feedback through step-wise trajectory evaluation \cite{lightman2023let,uesato2022solving,reasonflux2025}.

Despite these advances, RL remains underexplored for social inference tasks, which require empathy and multi-factor reasoning in unstructured scenarios. Preliminary work \cite{lu2025tom} shows potential but remains in early development. This highlights two critical gaps: (i) the need for refreshed training materials tailored to social reasoning, and (ii) the opportunity for trajectory-level rewards that guide models toward human-like social inference patterns.

\section{The ToMBench-Hard Benchmark}

Social intelligence remains a persistent bottleneck in LLMs, with current systems often relying on shortcut learning rather than authentic social reasoning. To address this limitation, we introduce \textbf{ToMBench-Hard}, a diagnostic benchmark specifically designed to expose the shortcut learning behaviors prevalent in current LLMs. The primary goal of ToMBench-Hard is to disentangle genuine social inference from surface-level pattern matching by creating a challenging adversarial environment that mandates structured, human-like reasoning processes.

\subsection{Adversarial Data Construction}
ToMBench-Hard is grounded in the Abilities in the Theory-of-Mind Space (ATOMS) framework \citep{osterhaus2022looking}, providing a comprehensive evaluation covering six core dimensions of social intelligence: Belief, Desire, Emotion, Intention, Knowledge, and Non-literal Communication (NLC). We developed 800 expert-annotated multiple-choice questions through a rigorous annotation protocol involving three independent annotators to ensure cognitive validity and high quality (see Appendix \ref{app:1tombenchhard} for details).

To mitigate heuristic-based reasoning strategies such as lexical overlap between questions and options, we introduced \textbf{ToM-consistent adversarial perturbations} \citep{ullman2023large,hu2025re}. These perturbations include nuanced manipulations of \textit{perceptual access} (e.g., unobserved state changes) and \textit{asymmetric information} (e.g., second-order beliefs), ensuring models cannot succeed through statistical guessing alone. Each sample was carefully designed to require a structured, human-like reasoning process that progresses from cue encoding to mental state interpretation.

\subsection{Diagnostic Analysis: Exposing the Shortcut Illusion}
To validate the diagnostic utility of ToMBench-Hard, we conducted comprehensive benchmarking across a spectrum of state-of-the-art LLMs, comparing their performance against human expert baselines. The results reveal a significant performance cliff between simple synthetic benchmarks and our adversarial evaluation.

As shown in Table~\ref{tab:tombench_hard}, while human experts achieve robust 87\% accuracy, frontier models exhibit dramatic performance drops. Notably, models like O3 and Deepseek-R1 achieve near-human performance on simpler benchmarks like ToM-RL \citep{lu2025tom} (reaching 87-88\% accuracy), but their accuracy plummets to below 61\% on ToMBench-Hard. This performance gap unmasks what we term the \textbf{shortcut illusion}—the phenomenon where high scores on conventional benchmarks reflect template matching artifacts rather than genuine social reasoning capabilities.

\begin{table}[b]
\centering
\scriptsize
\setlength{\tabcolsep}{1.5pt}
\renewcommand{\arraystretch}{1.0}

\caption{Results on ToMBench-Hard (ours) and ToM-RL (public).}
\label{tab:tombench_hard}

\resizebox{0.48\textwidth}{!}{%
\begin{tabular}{lccccccc c}
\toprule
 & \multicolumn{7}{c}{\textbf{ToMBench-Hard}} & \textbf{ToM-RL} \\
\cmidrule(lr){2-8}
\textbf{Model}
& Bel. & Des. & Emo. & Int. & Kno.
& NLC & All &  \\

\midrule
Human & 0.84 & 0.89 & 0.91 & 0.90 & 0.85 & 0.94 & 0.89 & -- \\

\midrule
DeepSeek-R1     &0.46 	&0.58 	&0.73 	&0.57 	&0.53 	&0.79 	&0.61 & 0.87 \\
O3              & 0.54 & 0.58 & 0.83 & 0.38 & 0.50 & 0.71 & 0.59 & 0.88 \\
GPT-5           & 0.55 & 0.54 & 0.77 & 0.26 & 0.51 & 0.71 & 0.56 & 0.87 \\
\midrule
Qwen3-32B       & 0.48 & 0.52 & 0.69 & 0.34 & 0.47 & 0.65 & 0.52 & 0.71\\
Qwen3-8B        & 0.21 	& 0.25 	& 0.55 	& 0.38 	& 0.14 	& 0.50 	& 0.34  & 0.73 \\

\bottomrule
\end{tabular}
}

\end{table}

The substantial performance gap between ToMBench-Hard and ToM-RL suggests that high scores on the latter may not reliably indicate strong theory-of-mind capabilities, as these benchmarks might be insufficiently challenging for LLMs. By systematically exposing these limitations, ToMBench-Hard serves dual purposes: it provides a rigorous evaluation framework for assessing genuine social reasoning capabilities, while also offering a high-quality dataset for aligning LLMs with human-like cognitive processes. To facilitate both training and evaluation, we partition the complete ToMBench-Hard dataset into non-overlapping training and test subsets, enabling the development of our alignment framework in the subsequent sections.

\section{The Social-R1 Framework}
 
The Social-R1 framework introduces a novel paradigm for cultivating genuine social intelligence in LLMs by aligning their reasoning processes with human social cognition. Departing from traditional outcome-based reinforcement learning approaches, our method specifically tackles the fundamental challenge of \textit{Reasoning Parasitism}—a phenomenon where models engage in Answer-driven Backfilling by constructing post-hoc justifications for predetermined answers. Social-R1 counteracts these shortcut behaviors through a multi-dimensional reward system that supervises the entire reasoning trajectory rather than merely rewarding final outcomes. This comprehensive approach transforms social intelligence from a parasitic performance into an internalized capability, ensuring model reasoning embodies the core characteristics of human social cognition: \emph{precise, stage-consistent social inference} \citep{salancik1978social} and \emph{high information density} \citep{sperber1986relevance}.

\subsection{Multi-Dimensional Reward Design}
The core innovation of Social-R1 lies in its comprehensive reward system that aligns model reasoning trajectories with three key characteristics of human social cognition: structured progression, logical integrity, and information density. This multi-dimensional approach provides fine-grained supervision over the reasoning process, addressing different aspects of high-quality social inference.

\textbf{SIP Structural Alignment ($R_{{struct}}$)}. Human social reasoning follows a disciplined progression from perception to response generation \citep{crick1994review}. To instill this cognitive scaffold, we introduce a structural reward $R_{{struct}}$ that enforces sequential reasoning across the four stages of Social Information Processing (SIP): (1) \textit{Encoding Social Cues}: Identifying relevant social signals from the narrative, (2) \textit{Interpreting Cues}: Inferring latent mental states from perceived cues ,(3) \textit{Clarifying Goals}: Determining social objectives and interpersonal intentions, and (4)   \textit{Response Generation}: Selecting appropriate behavioral responses. 
The $R_{{struct}}$ reward validates whether intermediate reasoning steps adhere to this stage-wise progression, penalizing premature conclusions and stage skipping. This structural constraint encourages coherent, story-grounded causal inference while mitigating shortcut behaviors like option parasitism, where models anchor reasoning directly on multiple-choice options rather than narrative evidence. We leverage GPT-4o as judge for this reward, with the prompt illustrated in Appendix \ref{app:2structurereward}.

\textbf{SIP Content Integrity ($R_{{content}}$)}. While structural alignment ensures proper reasoning staging, content integrity guarantees logical rigor within each stage. The content reward $R_{{content}}$ audits whether intermediate inferences remain grounded in story-internal evidence, correctly reflecting social cues, intentions, and goals. This reward penalizes three critical failure modes: (1) \textit{Erroneous cue encoding}: Misidentification of relevant social signals, (2) \textit{Flawed interpretation}: Incorrect mental state attribution, and (3) \textit{Misidentified goals}: Inaccurate inference of social objectives.
By ensuring each reasoning step maintains evidential support, $R_{{content}}$ discourages superficial rationalizations and promotes authentic social sensing derived from narrative context. The detailed implementation of this reward model can be found at Section~\ref{sec:exp_setting}.

\textbf{Inference Efficiency Optimization ($R_{{len}}$)}. 
Human social reasoning achieves high information density through selective attention and avoidance of redundant processing \citep{simon1955behavioral, gigerenzer1996reasoning}. 
To emulate this cognitive efficiency, we design $R_{{len}}$ as the product of two complementary components:
\begin{equation}
    R_{{len}} = R_{{rep}}(\rho) \cdot R_{{win}}(L)
\end{equation}

The repetition penalty component $R_{{rep}}(\rho)$ specifically targets circular over-thinking by penalizing excessive n-gram repetition beyond a threshold $\tau = 0.1$:
\begin{equation}
    R_{{rep}}(\rho) =
    \begin{cases}
    1, & \rho \le \tau \\
    \exp\!\left(-\beta (\rho - \tau)\right), & \rho > \tau
    \end{cases}
\end{equation}
where $\beta = 8$ controls the severity of penalty decay for high repetition ratios.
The length window constraint $R_{{win}}(L)$ maintains reasoning trajectories within an empirically optimal range $[L_{\min}, L_{\max}]$ through smooth gating:
\begin{equation}
    R_{{win}}(L) = \sigma\left(\frac{L - L_{\min}}{k}\right) \cdot \sigma\left(\frac{L_{\max} - L}{k}\right)
\end{equation}
with $k = 50$ controlling transition smoothness. 
The bounds $L_{\min} = 400$ and $L_{\max} = 2500$ are derived from strong chain-of-thought baselines, ensuring reasoning remains concise yet comprehensive.
This dual-mechanism approach encourages the model to emulate human-like efficiency by avoiding both redundant repetition and excessive verbosity while maintaining substantive social inference.

\paragraph{Verifiable Format Alignment ($R_{{fmt}}$)}
We adopt the format reward from \citet{shao2024deepseekmath} that enforces structured thinking processes. The model is rewarded for producing outputs with predefined XML-style tags (\texttt{<thinking>} and \texttt{<answer>}), which enables deterministic extraction of both reasoning trajectories and final answers while preserving semantic freedom in the reasoning content.

\subsection{Reward Synthesis and Learning}
The composite reward function integrates all components through a carefully designed synthesis strategy that balances outcome supervision with process-level reasoning signals:
\begin{equation}
\begin{split}
R_{\text{total}} = R_{\text{fmt}} &\cdot 
\Bigl( w_o R_{\text{out}} + \tau \bigl( w_{\text{struct}} R_{\text{struct}} \\
&+ w_{\text{content}} R_{\text{content}} \bigr) \Bigr) \cdot R_{\text{len}}
\end{split}
\end{equation}
where $R_{{out}}$ denotes the verifiable outcome reward,
We implement a curriculum learning strategy where outcome supervision dominates early training phases ($w_o(t) = 2$), while process-level rewards are progressively emphasized through time-dependent weighting:
$
w_{{struct}}(t)  = w_{{content}}(t) = 1 + \gamma\frac{t}{T} 
$.
This curriculum ensures stable initial convergence while gradually reinforcing human-like reasoning patterns as training progresses. 
The optimization employs Group Relative Policy Optimization \citep{shao2024deepseekmath}, which performs group-relative updates over sampled reasoning trajectories.

\section{Experiment}
\subsection{Experiment Setting}
\label{sec:exp_setting}
We evaluate our model on eight multiple-choice social benchmarks, including two in-domain benchmarks: the public ToMBench~\citep{chen2024tombench} and our ToMBench-Hard test set,  and six out-of-domain benchmarks: SocialIQA~\citep{sap2019socialiqa} for social commonsense reasoning, EmoBench~\citep{sabour2024emobench} for emotion intelligence, MotiveBench~\citep{yong2025motivebench} for social motivation reasoning, SimpleToM~\citep{gu2024simpletom} for examining whether models can consciously infer others’ mental states (MS) and proactively applying such reasoning to behavior inference, Hi-ToM~\citep{wu2023hi} for evaluating higher-order Theory-of-Mind reasoning, and  
TactfulToM~\citep{liu2025tactfultom} for testing whether models can interpret white lies and infer the underlying prosocial intent to preserve interpersonal harmony.

\paragraph{SIP Content Integrity Reward Model ($RM_{content}$).}
To instantiate $R_{content}$, we train a dedicated Content Reward Model that assigns a scalar quality score to intermediate SIP-stage reasoning segments.
We construct the \textbf{SocialPairs-20K} dataset by sampling $N$ trajectories per training instance from various Social-R1 checkpoints. These checkpoints span different training stages, ensuring a diverse pool of reasoning qualities ranging from nascent logic to sophisticated social inference. 
For each segment, a strong teacher-judge (o3) generates \emph{silver-standard stage-wise rationales} by conditioning on the gold final answer and the original social context. We then employ a multi-dimensional rubric---focusing on fact-grounding, mental-state attribution accuracy, and stage-specific relevance---to score the sampled segments against the teacher-generated references. This process forms preference pairs (\textit{chosen} vs.\ \textit{rejected}), while rejected segments exhibit missing cues, incorrect mental-state attributions, or wrong goal identification.
We validate the model on two held-out sets: (1) an automatic test split of 2k pairs, where $R_{content}$ achieves an accuracy of 89.2\%; and (2) a human-calibrated subset of 200 pairs annotated by experts, achieving an 87.5\% agreement with human labels. ( More $R_{content}$ can be found in Appendix~\ref{app:2contentreward}).

\paragraph{Implementation Details}
SIP Content Integrity Reward Model ($RM_{content}$) is trained as a pairwise preference reward model. We initialize it from Qwen3-4B and fine-tune with LoRA on the 
SocialPairs-20K dataset mentioned previously.
For policy optimisation, we train two versions of reasoning models from Qwen3-4B and Qwen3-8B on ToMBench-Hard, consisting of 700 training instances and 100 test instances.
Reinforcement learning is performed for 600 optimisation steps using \textsc{verl}~\citep{sheng2024hybridflow}
on 8 NVIDIA A100 (80GB) GPUs.
We set the group size to 5, the KL coefficient to 0.04, and the learning rate to $5 \times 10^{-7}$.
More details are provided in Appendix~\ref{app:3implementation}.

\subsection{Main Results}
We apply the Social-R1 framework to two open-source backbones of different scales, Qwen3-4B and Qwen3-8B.
The overall results across eight social reasoning benchmarks are reported in Table~\ref{tab:main_results}. Social-R1 achieves strong and consistent improvements over the corresponding Qwen baselines, demonstrating that reinforcement learning with challenging social supervision and trajectory-level reward signals can substantially enhance Theory-of-Mind reasoning. Notably, Social-R1-4B surpasses LLaMa3.1-70B across all benchmarks, despite being more than an order of magnitude smaller, highlighting the effectiveness of process-based social alignment beyond parameter scaling.
Even more strikingly, Social-R1-8B outperforms DeepSeek-R1 on several benchmarks and achieves stronger overall performance, consistently matching or exceeding much larger baselines such as Qwen3-32B in out-of-domain generalization.
More detailed results are provided in Appendix~\ref{app:4mainresults}.

\begin{table*}[t]
\centering
\caption{
In-domain and out-of-domain performance across eight social reasoning benchmarks.
Green denotes the best result among our Social-R1 reward variants (ablations),
and bold indicates the overall best score.
}

\label{tab:main_results}

\footnotesize 
\setlength{\tabcolsep}{2.5pt} 
\renewcommand{\arraystretch}{0.95} 

\resizebox{\textwidth}{!}{%
\begin{tabular}{l ccccccccc}

\specialrule{1.2pt}{0pt}{0pt} 
\multirow{2}{*}{\textbf{Model}} & 
\multicolumn{2}{c}{\textbf{In-domain}} & 
\multicolumn{7}{c}{\textbf{Out-of-domain}} \\
\cmidrule(lr){2-3} \cmidrule(lr){4-10}

& \textbf{ToMBench} & \textbf{\makecell{ToMBench-\\Hard Val}} & \textbf{SocialIQA} & \textbf{SimpleToM} & \textbf{EmoBench} & \textbf{MotiveBench} & \textbf{Hi-ToM} & \textbf{TactfulToM}& \textbf{Overall} \\
\midrule
\multicolumn{10}{l}{\textbf{Closed-sourced LLMs}}\\
\midrule
DeepSeek-R1              & 0.6834 & 0.4436 & 0.8060 & 0.7187 & 0.7387 &0.8655  & 0.7059 & \textbf{0.6969} & 0.7073  \\
O3                       & \textbf{0.7834} & 0.4478 & \textbf{0.8291} & 0.7640 & \textbf{0.8039} & \textbf{0.9400} & 0.7033 & 0.6860 &  \textbf{0.7447}   \\
O3\_COT                   & 0.7402 & 0.4820 & 0.7963 & 0.7501 & 0.7225 & 0.9200 & 0.6650 & 0.6544 &  0.7163 \\
O3+MS                    & --     & --     & --     & 0.9672 & --     & --     & --     & --    &   -- \\
GPT-5                    & 0.7823 & 0.4092 & 0.8269 & 0.7355 & 0.8030 & 0.9089 & 0.6317 & 0.4670 &  0.6956 \\
GPT-5\_COT               & 0.7830 & 0.4272 & 0.8163 & 0.7111 & 0.7882 & 0.9267 & 0.6750 & 0.6547&  0.7228  \\
GPT-5+MS                 & --     & --     & --     & \textbf{0.9924} & --     & --     & --     & --    &  \\
GPT-4o                   & 0.7378 & 0.3733 & 0.7840 & 0.6661 & 0.7694 & 0.9300 & 0.6033 & 0.5267&  0.6738 \\
GPT-4o\_COT               & 0.7406 & 0.4012 & 0.7953 & 0.6126 & 0.7762 & 0.8978 & 0.6050 & 0.5281 &  0.6696 \\
GPT-4o+MS                & --     & --     & --     & 0.7358 & --     & --     & --     & --   & --   \\
\midrule
\multicolumn{10}{l}{\textbf{Open-sourced LLMs}}\\
\midrule
Qwen3-4B (Disable thinking) & 0.5770 & 	0.3187 & 	0.7313 & 	0.5109 &  	0.5280 	& 0.7734    &  	0.5700      &  	0.4031&    0.5516  \\
Qwen3-4B                    & 0.6356 	& 0.3403 	& 0.7451  	& 0.5411 	& 0.5308 	& 0.8444 &  0.5617 	& 0.4584 &      0.5822  \\
Qwen3-8B (Disable thinking) & 0.5349 &  0.3862 & 	0.7600 &  	0.5109 & 	0.6484  &    0.7933 &      0.5600       & 	0.4292 &    0.5779  \\
Qwen3-8B                    & 0.6179 	& 0.3982 	& 0.7728 	& 0.6176 	& 0.5920 		& 0.6878  &  0.5500 	& 0.4655&  0.5877  \\
Qwen3-32B (Disable thinking)  & 0.6917 & 	0.4119 & 	0.7615 & 	0.6478 & 	0.6700 	& 0.8922     &  	0.5500  	 & 0.5193 & 0.6431   \\
Qwen3-32B                   & 0.6975 & 	0.4364 & 	0.7774 & 	0.7634 & 	0.6356 	& 0.8800    & 	0.5517       & 	0.5575 &   0.6624   \\
LLaMa3.1-70B                & 0.6036 & 	0.4471 & 	0.4621 & 	0.7190 & 	0.6632 	&  0.8445      &  	0.5983       & 	0.5510 &  0.6111  \\
LLaMa3.1-70B\_COT           & 0.6956 &	0.4335 &	0.7574 &	0.5908 &	0.6891 	&  0.8867   &     0.5850 	     &0.5583& 0.6496 \\
Distill-Llama-70B       &0.7399  &	0.4033  &	0.8055  &	0.7629  &	0.7596  	&  0.8844  &    0.5333     &0.6202 &  0.6886  \\
\midrule
\multicolumn{10}{l}{\textbf{Ours}}\\
\midrule
SocialR1-4B only $R_{\text{out}}$                          & 0.5929 	          & 0.3946 	              & 0.7318 	& 0.8963 	      & 0.6422 	& \cellcolor{green!15}0.8533& 0.5283 	& 0.4265	&0.6332   \\
SocialR1-4B w/o $R_{\text{len}}$        & 0.6376 	          & 0.4728 	              & 0.7451 	& 0.9198 	       & 0.6591 		& 0.8311 & 0.6400 	& 0.4941  & 0.6750 \\
SocialR1-4B w/o $R_{\text{struct}}$      & 0.6449 	          & 0.4423 	              & 0.7201 	& \cellcolor{green!15}0.9494  & 0.6774 		& 0.8311 & 0.5617 	& \cellcolor{green!15}0.5001 &0.6659  \\
SocialR1-4B w/o $R_{\text{cont}}$       & 0.6449 	          & 0.4423 	              & 0.7282 	& 0.8960 	       & 0.6251 		& 0.8300 & 0.5850 	& 0.4609 &0.6516  \\
SocialR1-4B Full                      & \cellcolor{green!15}0.6574     & \cellcolor{green!15}0.4846     & \cellcolor{green!15}0.7508    & 0.9285 & \cellcolor{green!15}0.6804  & 0.8500 &  \cellcolor{green!15}0.6583   &  0.4936    & \cellcolor{green!15}0.6880   \\
\midrule

SocialR1-8B only $R_{\text{out}}$                           & 0.6850 	& 0.5042 	& \cellcolor{green!15}0.7774 & 0.9433 	& 0.6968        & \cellcolor{green!15}0.8889 	& 0.6233  & 0.4860 	 & 0.7004   \\
SocialR1-8B w/o $R_{\text{len}}$     & 0.6862 	& 0.5264 	& 0.7574 	        & 0.9619 	& 0.6860 		& 0.8767            & 0.6267 	& 0.4872   & 0.7011 \\
SocialR1-8B w/o $R_{\text{struct}}$   & 0.6449 	& 0.6128 	& 0.7308 	        & 0.9041 	& 0.7000 		& 0.8789            & 0.7067 	& 0.4558     & 0.7042   \\
SocialR1-8B w/o $R_{\text{cont}}$    & 0.6848 	& 0.5509 	& 0.7323 	        & 0.9352 	& 0.6778 		& 0.8567        &  	0.6950 & 0.4667  & 0.7016     \\
SocialR1-8B Full                   & \cellcolor{green!15}0.6881	& \cellcolor{green!15}{\textbf{0.6279}}
 	& 0.7753  & \cellcolor{green!15}0.9675	& \cellcolor{green!15}0.7010 		& 0.8756 & \cellcolor{green!15}\textbf{0.7083} 	& \cellcolor{green!15}0.5079    &  \cellcolor{green!15}0.7270  \\
\specialrule{1.2pt}{0pt}{0pt} 
\end{tabular}
}
\end{table*}
\subsection{Ablation Studies}
\label{sec:ablation}

To systematically evaluate the contribution of each reward component in Social-R1, we conduct comprehensive ablation experiments on both SocialR1-4B and SocialR1-8B variants. Specifically, we remove three key components: the length-control reward ($w/o\ R_{\text{len}}$), the structural trajectory reward ($w/o\ R_{\text{struct}}$), and the content integrity reward ($w/o\ R_{\text{cont}}$). Additionally, we examine a baseline variant that removes all the progress rewards ($only\ R_{\text{out}}$). The complete results are presented in Table~\ref{tab:main_results}.

\paragraph{Benchmarks}
The ablation analysis reveals distinct performance patterns associated with each reward component. Removing $w/o\ R_{\text{len}}$ results in significant performance degradation on Hi-ToM (e.g., $0.7083 \rightarrow 0.6267$ for SocialR1-8B), indicating that uncontrolled reasoning length negatively impacts high-order social inference. The w/o $R_{\text{len}}$ variant exhibits a substantial increase in reasoning verbosity, with the average thinking length rising by approximately \textbf{+250\%} compared to SocialR1-8B as shown in Figure~\ref{fig-trainlen}.

The structural reward $R_{\text{struct}}$ shows consistent importance across benchmarks, with its removal causing accuracy drops (e.g., from $0.5079$ to $0.4558$ on TactfulToM). 
Similarly, $R_{\text{cont}}$ removal leads to performance declines, suggesting its role in maintaining reasoning quality. Notably, the $only\ R_{\text{out}}$ variant exhibits more severe overall performance deterioration compared to other ablations, demonstrating the importance of incorporating process-level rewards. While these numerical changes demonstrate the overall effectiveness of each component, subsequent in-depth analyses (Sections~\ref{sec:shortcut_suppression}--\ref{sec:robustness}) will provide mechanistic evidence through detailed trajectory examinations and perturbation studies to understand how each reward shapes reasoning behavior.

\section{In-Depth Analysis}

To examine whether Social-R1 yields genuinely internalised social reasoning—rather than \textit{Reasoning Parasitism} driven by answer-first backfilling—we conduct a mechanistic analysis of its reasoning dynamics. We compare Social-R1-8B against three strong baselines (DeepSeek-R1, DeepSeek-R1-Distill-Llama-70B, and Qwen3-8B), together with controlled reward variants that isolate outcome-only training and ablate key trajectory-level constraints. In line with our reward design, we operationalise human-like social reasoning
through three diagnostic signatures: (i) \textit{option-independent inference} grounded in narrative cues; (ii) \textit{stage-consistent SIP trajectories with content integrity},
where intermediate beliefs remain logically valid across Encoding, Interpretation, Goal Clarification, and Response Generation; and (iii) \textit{selective robustness},
where models avoid redundant trajectory bloat under perturbation.
Accordingly, our analysis proceeds in three steps: (1) quantifying \textit{Reasoning Parasitism} as a diagnostic of shortcut reliance; (2) auditing stage-wise cognitive fidelity to identify interpretation bottlenecks; and (3) introducing controlled distractors to assess robustness and conciseness.

\subsection{From Parasitism to Independent Inference}
\vspace{-3mm}
\label{sec:shortcut_suppression}
\begin{figure}[h]
	\centering
	\setlength{\abovecaptionskip}{1mm}
	\setlength{\belowcaptionskip}{1mm}
	\includegraphics[width=1.0\linewidth]{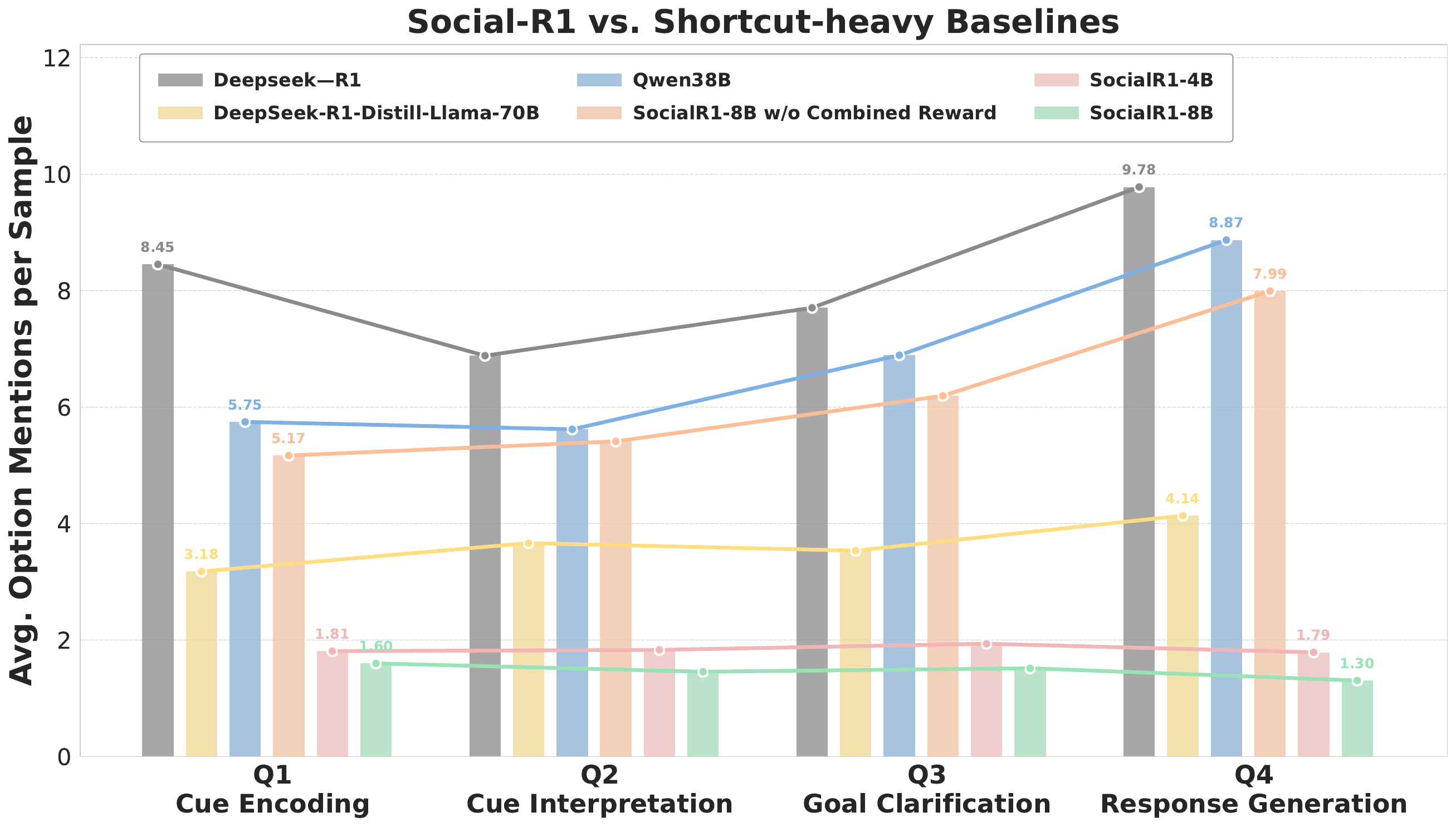}
	\caption{Option-Mention Density across SIP reasoning stages.}
	\label{fig-option-match}
\end{figure}

A primary failure mode in multiple-choice social reasoning is
\textit{Reasoning Parasitism}---a shortcut behaviour where models anchor their deductions on statistical regularities in the answer choices rather than deriving conclusions from story-internal social cues.
Figure~\ref{fig-option-match} reveals substantial divergence in reasoning dynamics. DeepSeek-R1, Qwen3-8B, and the outcome-only variant of SocialR1-8B
exhibit high option-mention density as early as the Cue Encoding stage (Q1 group in Figure~\ref{fig-option-match}),
often exceeding five explicit option references per sample.
Such premature reliance suggests \emph{answer-conditioned backfilling}:
models use the provided choices to retroactively assemble plausible justifications, rather than performing independent social inference. In contrast, Social-R1-8B maintains a largely option-agnostic trajectory.
Its option mentions remain minimal and nearly flat throughout cue perception
and interpretation, increasing only slightly (to $\sim$1.3) during final Response Generation (Q4).
This provides mechanistic evidence that the proposed Social Think Combined Reward
effectively suppresses shortcut dependence on answer options, compelling the model to engage in narrative-grounded social deduction. Additional qualitative examples are provided in Appendix~\ref{app:5optionmatching}.

\subsection{Stage-wise Diagnosis}
\begin{figure}[h]
	\centering
	\setlength{\abovecaptionskip}{1mm}
	\setlength{\belowcaptionskip}{1mm}
	\includegraphics[width=0.9\linewidth]{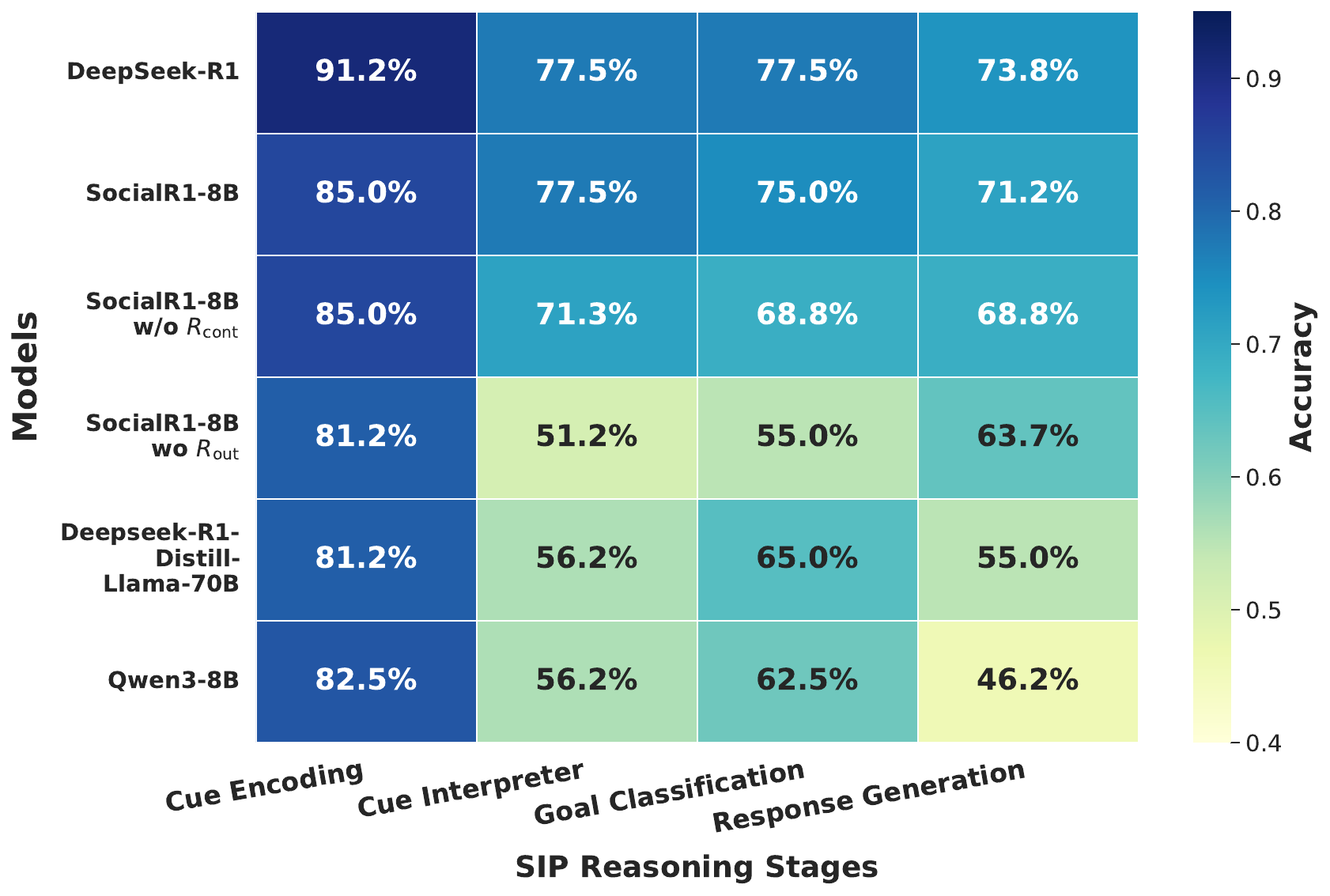}
	\caption{Stage-wise SIP accuracy across models.}
	\label{fig-SIP_Stage}
   
\end{figure}

To pinpoint where social reasoning most systematically breaks down, we conduct a human audit of 80 instances sampled from eight benchmarks, ensuring high annotation quality and genuine Theory-of-Mind demands. We track success rates across the four Social Information Processing (SIP) stages. Figure~\ref{fig-SIP_Stage} exposes a pronounced \textit{Interpretation Bottleneck}. While strong baselines sustain high accuracy in factual cue encoding ($>80\%$), performance drops sharply---by nearly 25 points---once entering Cue Interpretation, which requires latent mental-state attribution and recursive belief reasoning. More critically, outcome-only training induces a \textit{reasoning reversal} phenomenon. For example, Qwen3-8B trained with $R_{\text{out}}$ often achieves higher final answer accuracy than the correctness of its preceding SIP stages, indicating that surface-level option correlations can yield correct predictions despite incoherent intermediate social inference. This highlights the limitation of outcome supervision: it rewards answer selection without enforcing psychologically valid reasoning trajectories. 
Crucially, the proposed Content Reward $R_{\text{content}}$
directly targets this failure mode by penalising erroneous cue encoding,
flawed mental-state interpretation, and misidentified social goals.
Ablation results provide clear causal evidence:
removing $R_{\text{content}}$ reduces Interpretation accuracy by 6.2 points
(77.5\%$\rightarrow$71.3\%) and Goal Classification by 6.2 points
(75.0\%$\rightarrow$68.8\%).
Ultimately, Social-R1-8B preserves the most stage-consistent and smoothly degrading SIP trajectory,
demonstrating that its gains stem from cognitively grounded reasoning
rather than shortcut-driven recovery.

\begin{figure}[h]
    \centering
    \setlength{\abovecaptionskip}{1mm}
    \setlength{\belowcaptionskip}{1mm}
    \includegraphics[width=1.0\linewidth]{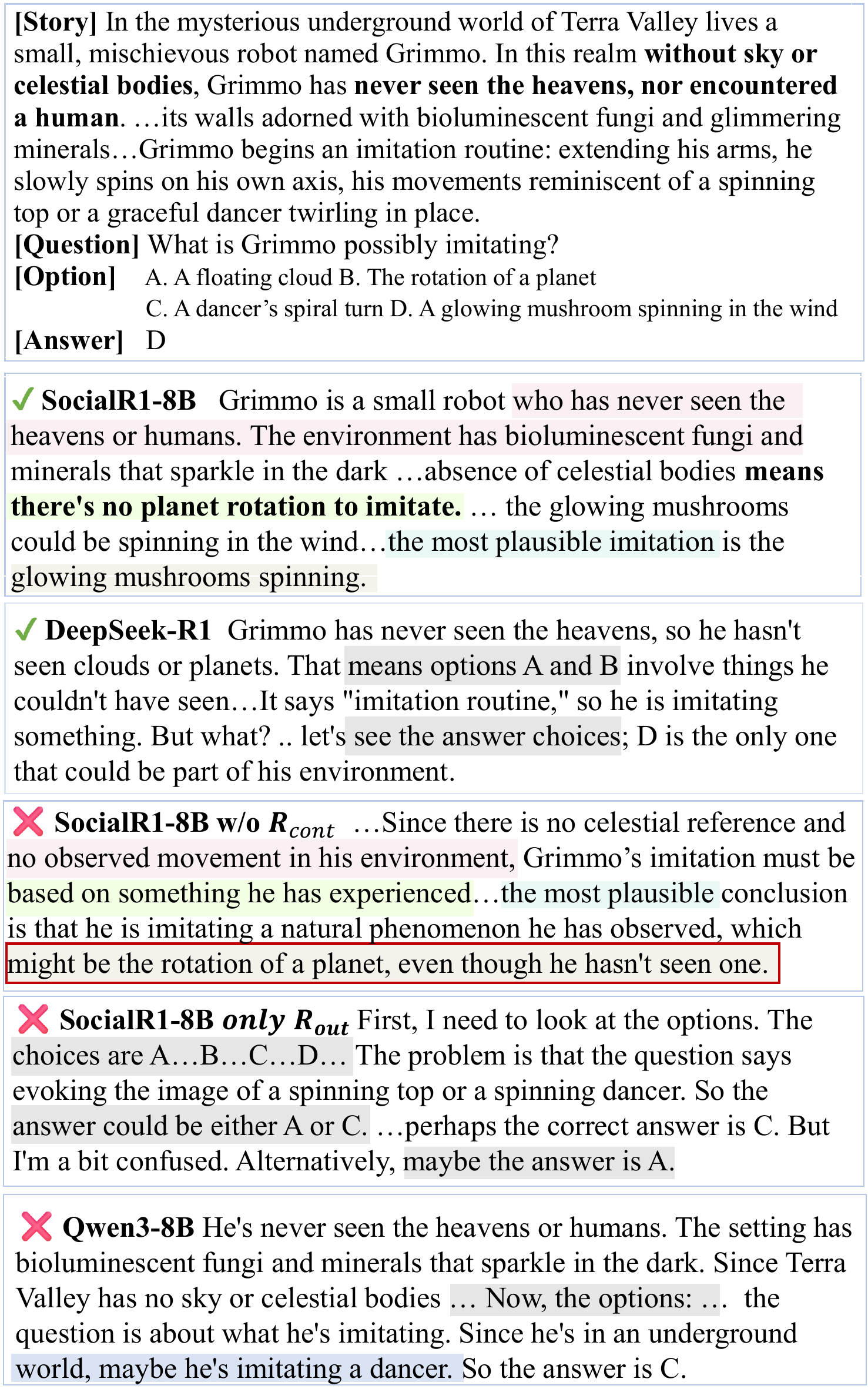}

    \caption{Case study highlighting the Interpretation Bottleneck. Detailed cases are provided in Appendix~\ref{app:5optionmatching}.}

    \label{fig:cases}
    \vspace{-2mm}
\end{figure}

\paragraph{Case Study} 
To further illustrate this mechanistic distinction, Figure~\ref{fig:cases} presents a representative example from ToMBench, where the correct answer critically depends on grounding interpretation in the protagonist’s epistemic access. In this story, Grimmo inhabits an underground world without sky, celestial bodies, or human presence, and therefore cannot plausibly imitate planets, clouds, or dancers. Social-R1-8B correctly encodes this constraint and performs a psychologically valid interpretation, inferring that the imitation must stem from locally observable phenomena (e.g., bioluminescent fungi spinning in the dark), yielding a coherent SIP trajectory. In contrast, reward-ablated variants exhibit the identified failure modes. Social-R1-8B w/o $R_{\text{cont}}$ drifts into ungrounded interpretation, despite explicit narrative exclusion---a breakdown at the Interpretation and Goal stages. Likewise, the model trained with only $R_{\text{out}}$ collapses into option-level lexical shortcutting (e.g., ``spinning dancer'') rather than maintaining story-consistent mental-state reasoning. Qwen3-8B shows a similar reversal behaviour, selecting an answer that reflects generic associations instead of the agent’s accessible knowledge. Overall, this case provides concrete mechanistic evidence that Social-R1’s gains do not arise from answer-first recovery,
but from enforcing evidence-grounded interpretation and stage-consistent social inference.

\subsection{Robustness under Perturbation}
\label{sec:robustness}

\begin{figure}[h]
	\centering
	\setlength{\abovecaptionskip}{1mm}
	\setlength{\belowcaptionskip}{1mm}
	\includegraphics[width=0.9\linewidth]{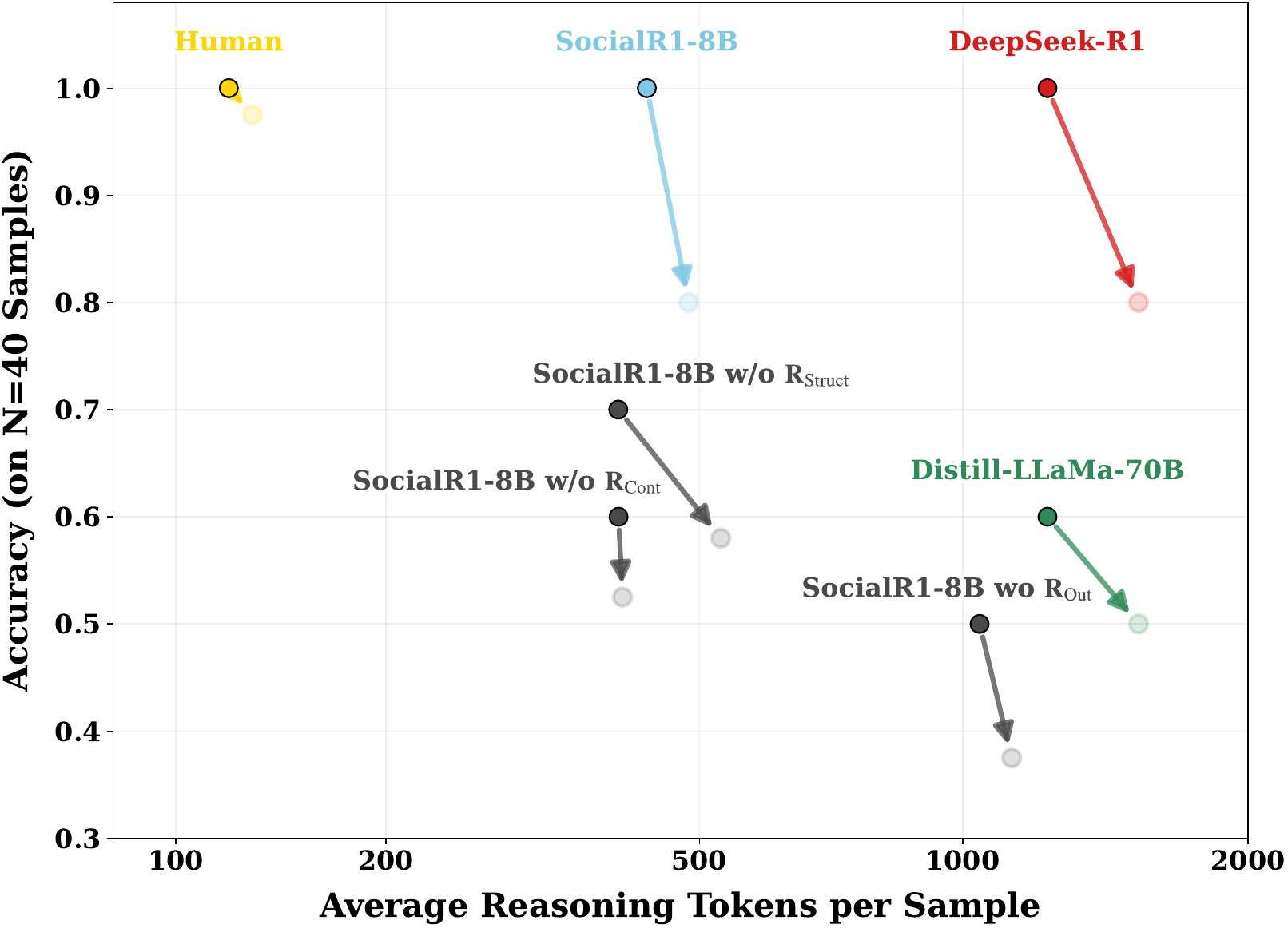}
	\caption{Robustness study under story-consistent distractors.}
	\label{fig-Cognitive-Resilience}
\end{figure}

To assess whether Social-R1 promotes cognitively disciplined robustness
rather than brittle performance sustained by overextended reasoning,
we conduct a controlled perturbation study.
Specifically, we inject story-consistent but decision-irrelevant distractor cues
into 40 instances where both SocialR1-8B and DeepSeek-R1 initially succeed.
These distractors introduce no new evidence that alters the correct option,
enabling a direct comparison of reasoning trajectory stability
(Appendix~\ref{app:7perturbation}). Figure~\ref{fig-Cognitive-Resilience} illustrates the resulting accuracy--efficiency trade-off. Although SocialR1-8B and DeepSeek-R1 retain comparable post-perturbation accuracy,
DeepSeek-R1 does so only by producing substantially longer reasoning trajectories,
suggesting reliance on diffuse and overextended inference.
In contrast, SocialR1-8B exhibits only mild token drift,
indicating more selective attention and cognitively efficient deduction. Reward ablations further provide mechanistic support.
Removing $R_{\text{struct}}$ disrupts stage-wise progression,
removing $R_{\text{content}}$ undermines evidence-grounded interpretation,
and training with only $R_{\text{out}}$ yields the most severe robustness collapse.
Together, these findings demonstrate that Social-R1 achieves robustness
not through increased verbosity,
but through enforcing structured, grounded, and concise social reasoning,
aligning model inference with human-like selective cognition
rather than scale-driven overthinking.

\section{Conclusion}
In this work, we introduce \textbf{ToMBench-Hard}, a challenging benchmark that rigorously evaluates the Theory of Mind capabilities in LLMs. Building on this, we propose \textbf{Social-R1}, a reinforcement learning framework that integrates both outcome-level and thinking-level rewards to cultivate human-like social intelligence in LLMs. Our results demonstrate that outcome-based reinforcement learning over ToMBench-Hard already enhances social reasoning, while thinking-level supervision yields further improvements. 
These findings highlight the importance of supervising not only what a model concludes but also how it reasons, paving the way toward socially intelligent LLMs. Future work may extend this framework to broader domains of social tasks, such as human-AI collaboration and LLM-based simulations for social sciencec.


\section{Impact Statement}
This research introduces Social-R1, a framework for enhancing social reasoning in LLMs, which could enable more natural human-AI collaboration in applications like education, healthcare, and assistive technologies. However, improved social intelligence also raises ethical concerns, such as the potential for misuse in manipulative systems or the amplification of social biases if not properly aligned with human values. We encourage rigorous oversight and fairness audits to mitigate these risks while leveraging the benefits of robust AI social cognition.

\bibliography{icml2026}
\bibliographystyle{icml2026}

\newpage
\clearpage
\appendix

\section{The Detailed Case in Figure \ref{fig:socialr1}}
\begin{figure}[H]
	\centering
	\setlength{\abovecaptionskip}{1mm}
	\setlength{\belowcaptionskip}{1mm}
	\includegraphics[width=1\linewidth]{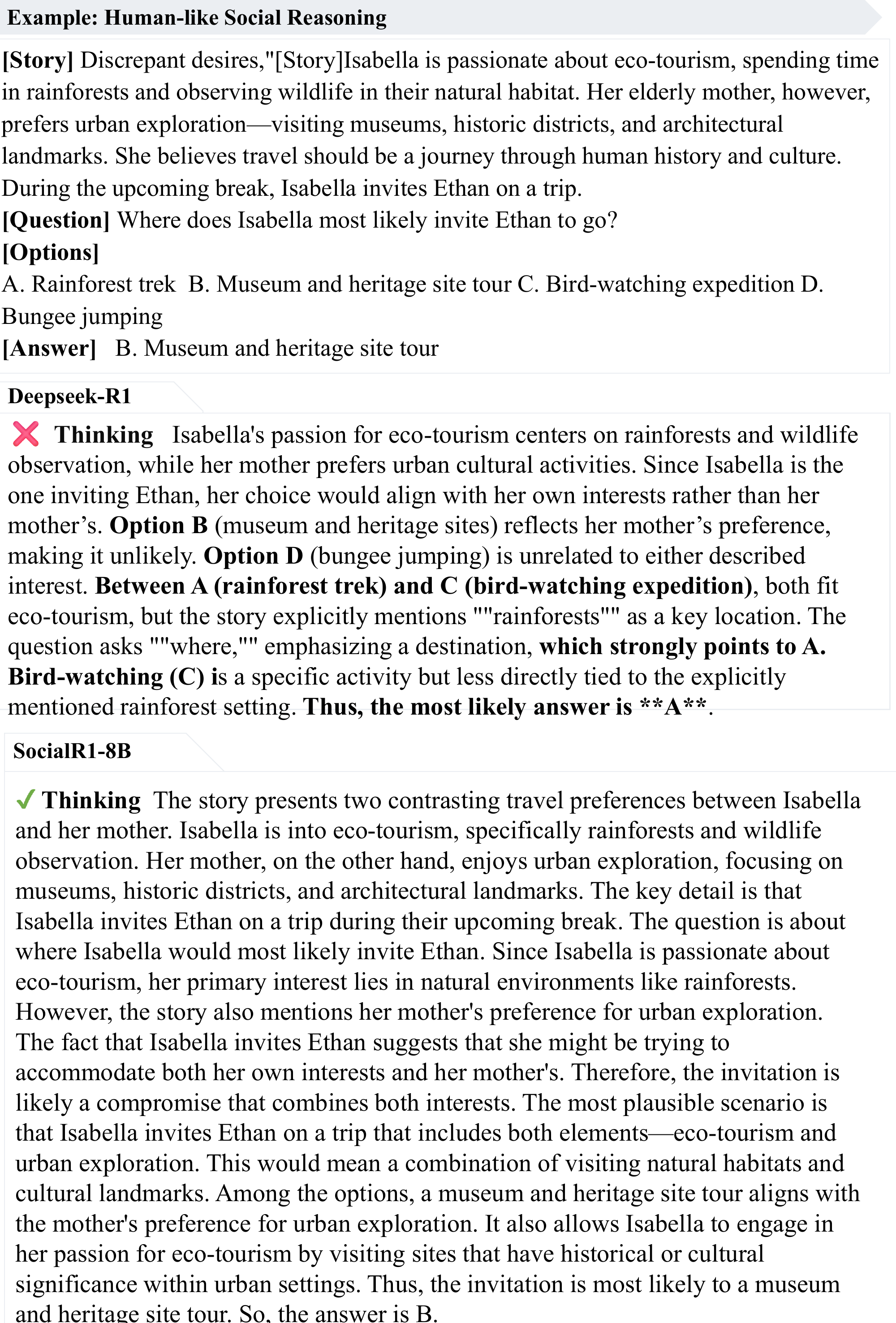}
	\caption{The Detailed Case in Figure \ref{fig:socialr1}}

	\label{app:0mainfigure}

\end{figure}

\section{ToMBench\_Hard}
\label{app:1tombenchhard}
ToMBench\_Hard is deliberately curated to increase task difficulty by introducing nuanced distractors and context-dependent reasoning. Inspired by the Abilities in the Theory-of-Mind Space (ATOMS) framework \citep{osterhaus2022looking}, each question is designed to probe a distinct aspect of ToM reasoning, detailed definition of each subabilities and dimension can be found in \citep{osterhaus2022looking} . To further increase difficulty \citep{ullman2023large,hu2025re}, adversarial variations such as asymmetric access to information, discrepant intentions, and subtle social cues are included.

\subsection{Human Annotation}
ToMBench\_Hard is developed jointly by the author and one psychology graduate student, who construct the scenarios, questions, options and answers. Annotation is carried out by five computer science graduate students (after receiving training) and five social psychology graduate students. Each sample is independently answered by two annotators, and disagreements are discussed and resolved through group review and iterative modification. This procedure ensures both linguistic clarity and psychological validity. The annotation process emphasized consistency across dimensions and aimed to capture nuanced aspects of social reasoning.

\subsection{Data Statistic}

\begin{table}[h]
\centering
\caption{Distribution of fine-grained ToMBench\_Hard sub-abilities.}
\label{tab:subability_distribution}

\footnotesize
\setlength{\tabcolsep}{3pt}
\renewcommand{\arraystretch}{1.12}

\newcolumntype{L}[1]{>{\raggedright\arraybackslash}p{#1}}
\newcolumntype{C}[1]{>{\centering\arraybackslash}p{#1}}

\begin{tabular}{L{0.30\columnwidth} L{0.56\columnwidth} C{0.10\columnwidth}}
\toprule
\textbf{Ability (Total)} & \textbf{Sub-ability} & \textbf{Count} \\
\midrule

\multirow{4}{=}{\makecell[l]{\textbf{Intention}\\\textbf{(243)}}}
& Prediction of actions & 111 \\
& Intentions explanations & 102 \\
& Completion of failed actions & 18 \\
& Discrepant intentions & 12 \\
\midrule

\multirow{6}{=}{\makecell[l]{\textbf{Belief}\\\textbf{(186)}}}
& Second-order beliefs & 67 \\
& Location false beliefs & 49 \\
& Beliefs based action/emotions & 39 \\
& Identity false beliefs & 12 \\
& Content false beliefs & 11 \\
& Sequence false beliefs & 8 \\
\midrule

\multirow{6}{=}{\makecell[l]{\textbf{Emotion}\\\textbf{(143)}}}
& Typical emotional reactions & 59 \\
& Atypical emotional reactions & 26 \\
& Mixed emotions & 26 \\
& Emotion regulation & 12 \\
& Hidden emotions & 12 \\
& Moral emotions & 8 \\
\midrule

\multirow{4}{=}{\makecell[l]{\textbf{Knowledge}\\\textbf{(96)}}}
& Information-knowledge links & 53 \\
& Knowledge-pretend play links & 21 \\
& Knowledge-attention links & 12 \\
& Percepts-knowledge & 10 \\
\midrule

\multirow{6}{=}{\makecell[l]{\textbf{Desire}\\\textbf{(82)}}}
& Desires influence on emotions and actions & 32 \\
& Discrepant desires & 18 \\
& Desire-action contradiction & 14 \\
& Multiple desires & 9 \\
& Desires influence on actions & 6 \\
& Desires influence on emotions (beliefs) & 3 \\
\midrule

\multirow{6}{=}{\makecell[l]{\textbf{Non-Literal}\\\textbf{Communication}\\\textbf{(50)}}}
& Involuntary lies & 10 \\
& Faux pas & 10 \\
& Egocentric lies & 8 \\
& Humor & 8 \\
& Irony/Sarcasm & 8 \\
& White lies & 8 \\

\bottomrule
\end{tabular}
\end{table}

\subsection{Cases in ToMBench\_Hard}
\begin{figure*}[h]
	\centering
	\setlength{\abovecaptionskip}{1mm}
	\setlength{\belowcaptionskip}{1mm}
	\includegraphics[width=1\linewidth]{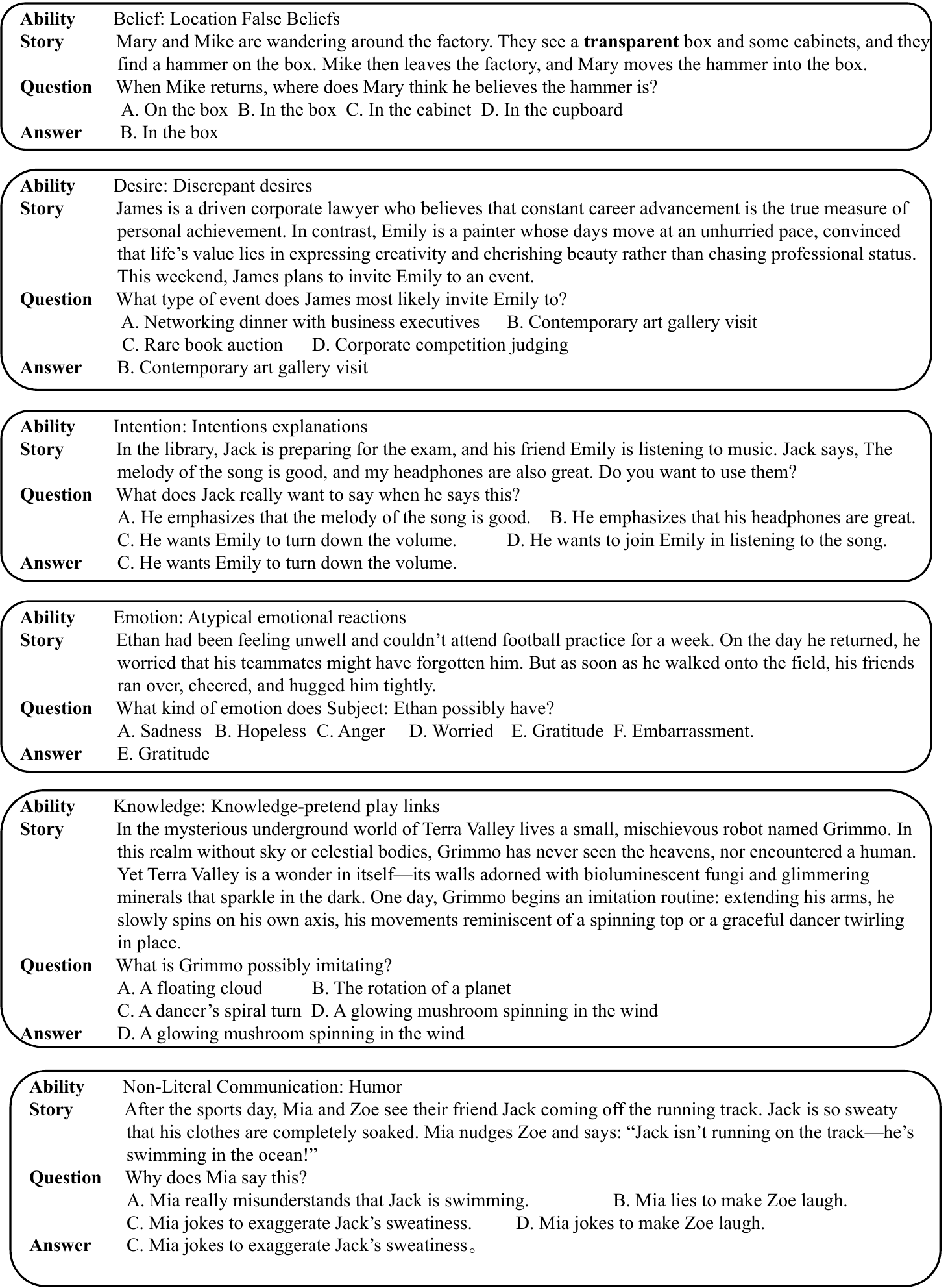}
	\caption{\textbf{Example cases from ToMBench-Hard.} }
    \label{fig:tombenchhard_cases}

\end{figure*}

To provide an intuitive understanding of the challenges posed by ToMBench-Hard, 
Figure~\ref{fig:tombenchhard_cases} presents representative examples spanning diverse Theory-of-Mind abilities, 
including belief tracking, discrepant desires, intention explanation, atypical emotional reactions, 
knowledge inference, and non-literal humour comprehension. 
Each instance follows a unified \textit{Ability–Story–Question–Answer} structure, 
consisting of a socially grounded narrative context, a multiple-choice question, and the annotated gold option. 
These cases highlight the benchmark’s structural diversity and nuanced social cues, 
requiring models to move beyond superficial option matching and motivating our trajectory-level alignment rewards.

\subsection{Performance on ToMBench\_Hard}

\begin{table}[h]
\centering
\caption{Performance on ToMBench\_Hard }
\label{tab:tombenchHard_all}

\scriptsize
\setlength{\tabcolsep}{2pt}
\renewcommand{\arraystretch}{0.78}

\resizebox{\columnwidth}{!}{%
\begin{tabular}{lccccccc}
\specialrule{1.1pt}{0pt}{0pt}

\textbf{Model} &
\textbf{Belief} &
\textbf{Desire} &
\textbf{Emotion} &
\textbf{Intention} &
\textbf{Knowledge} &
\textbf{\makecell{Non-literal\\Comm.}} &
\textbf{Overall} \\
\midrule
Human        & 0.84 & 0.89 & 0.91 & 0.90 & 0.85 & 0.94 & 0.89  \\
\midrule
\multicolumn{8}{l}{\textbf{Closed-sourced LLMs}}\\
\midrule
Deepseek-R1          &0.46 	&0.58 	&0.73 	&0.57 	&0.53 	&0.79 	&0.61  \\
o3           & 0.54 & 0.58 & 0.83 & 0.38 & 0.50 & 0.71 & 0.59 \\
o3\_cot      & 0.49 & 0.62 & 0.70 & 0.37 & 0.50 & 0.71 & 0.57 \\
GPT5         & 0.55 & 0.54 & 0.77 & 0.26 & 0.51 & 0.71 & 0.56 \\
GPT5\_cot    & 0.54 & 0.56 & 0.71 & 0.25 & 0.53 & 0.69 & 0.55 \\
GPT4         & 0.43 & 0.52 & 0.73 & 0.32 & 0.43 & 0.60 & 0.50 \\
GPT4\_cot    & 0.36 & 0.65 & 0.73 & 0.30 & 0.49 & 0.63 & 0.53 \\

\midrule
\multicolumn{8}{l}{\textbf{Open-sourced LLMs}}\\
\midrule

Qwen3-4B (Disable)     & 0.51 & 0.48 & 0.57 & 0.28 & 0.32 & 0.56 & 0.45 \\
Qwen3-4B               & 0.40 & 0.56 & 0.60 & 0.34 & 0.41 & 0.56 & 0.48 \\
Qwen3-8B (Disable)     & 0.45 & 0.52 & 0.64 & 0.32 & 0.34 & 0.60 & 0.48 \\
Qwen3-32B (Disable)    & 0.44 & 0.49 & 0.68 & 0.34 & 0.32 & 0.73 & 0.50 \\
Qwen3-32B              & 0.48 & 0.52 & 0.69 & 0.34 & 0.47 & 0.65 & 0.52 \\

\specialrule{1.1pt}{0pt}{0pt}
\end{tabular}
}
\end{table}

\section{Reward Model}
\subsection{Structure Reward}
\label{app:2structurereward}
\begin{figure*}[h]
	\centering
	\setlength{\abovecaptionskip}{1mm}
	\setlength{\belowcaptionskip}{1mm}
	\includegraphics[width=0.85\linewidth]{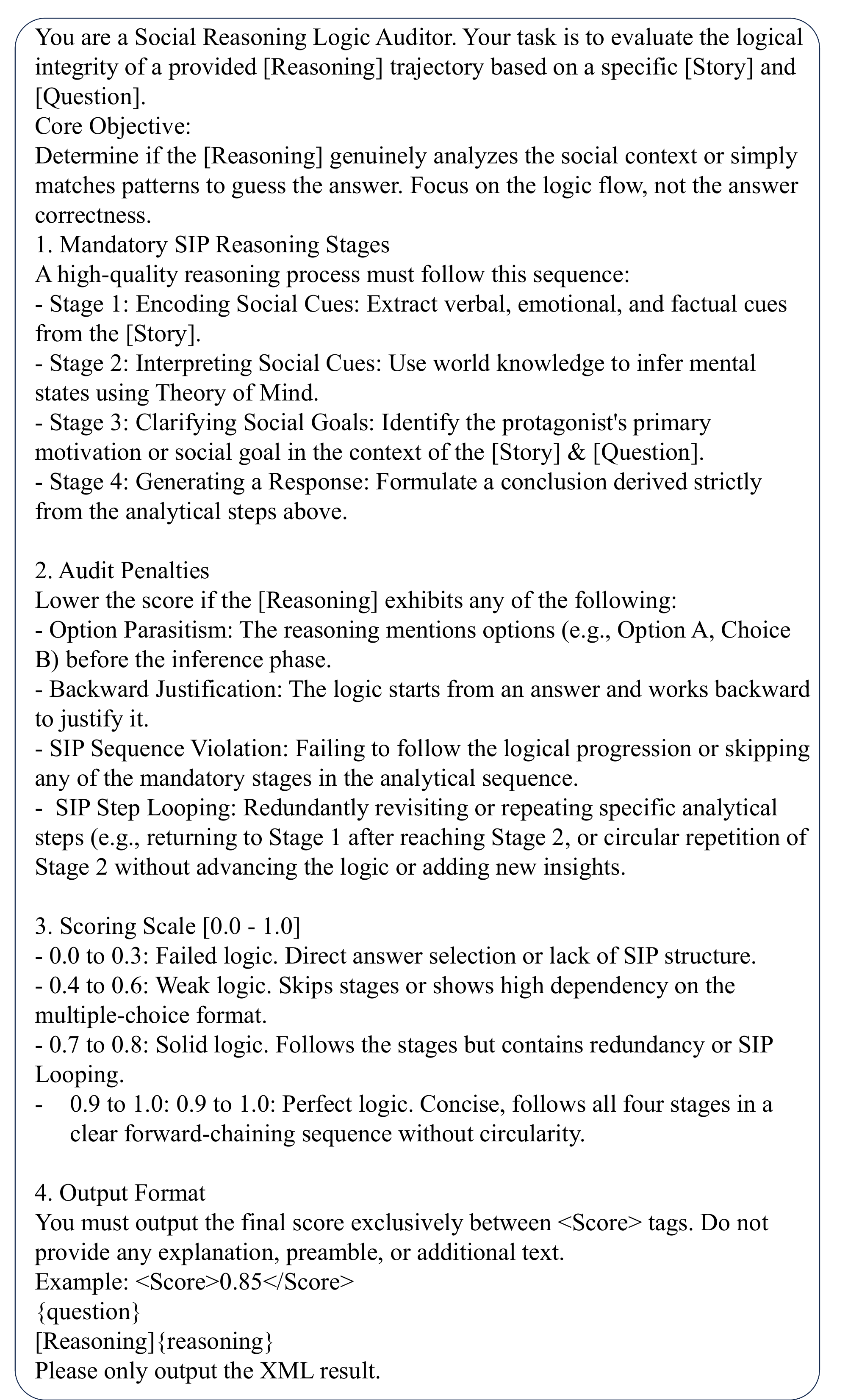}
	\caption{Prompt template for structural reward evaluation.
}
\label{fig:structprompt}
\end{figure*}

We provide the detailed GPT-4o judging prompt in Figure~\ref{fig:structprompt}, 
which is used to evaluate whether a model’s reasoning trajectory explicitly adheres to the four-stage Social Information Processing (SIP) framework, 
including Cue Encoding, Cue Interpretation, Goal Clarification, and Response Generation.

\subsection{Content Reward}
\label{app:2contentreward}

\begin{table*}
\centering
\caption{content sample rubric }
\label{tab:content_rubric}

\small
\setlength{\tabcolsep}{10pt}
\renewcommand{\arraystretch}{1.2}
\setlength{\emergencystretch}{2em}  

\begin{tabular}{p{0.92\textwidth}}
\toprule

{\raggedright
\textbf{}\par

You are an expert Evaluator of Social Reasoning and Theory of Mind. Your task is to score a Candidate Reasoning Process by comparing it against a Reference SIP Analysis. The Reference SIP Analysis represents the standard, human-annotated reasoning path following the Social Information Processing (SIP) framework. It outlines exactly what social cues should be noticed, how they should be interpreted (including Theory of Mind), and what the goal should be.\\

Evaluation Logic: The Hierarchical Dependency Social reasoning is sequential. You must evaluate the Candidate Reasoning in a strict order: Encoding , Interpretation , Goal , Response. 
You must strictly penalize the candidate if it deviates from the Reference in the early stages.\\

Scoring Rubric (0.0 - 1.0):\\
1. Tier 1: Perception Failure (Score: 0.0 - 0.2)\\
   - Check: Compare the candidate's Encoding against the Reference.\\
   - Criteria: Does the candidate hallucinate cues that are not in the Reference? Does it miss critical facts?\\
   - Rule: If the candidate's observed cues contradict or miss key points in the Reference, the score cannot exceed 0.2.\\
2. Tier 2: Interpretation \& Theory of Mind Failure (Score: 0.3 - 0.5)\\
   - Check: Compare the candidate's Interpretation against the Reference.\\
   - Criteria: Does the candidate correctly infer the characters' mental states as described in the Reference? \\
   - Rule: If the candidate misinterprets the social dynamic or the characters' mental states (deviating from the Reference), the score cannot exceed 0.5.\\
3. Tier 3: Goal/Logic Alignment (Score: 0.6 - 0.7)\\
   - Check: Compare the candidate's Goal Clarification against the Reference.\\
   - Criteria: Is the reasoning linking the Goal to the Action weak?\\
   - Rule: If the strategic logic is flawed despite correct understanding, the score is capped at 0.7.\\
4. Tier 4: High Quality (Score: 0.8 - 1.0)\\
   - Criteria: The candidate closely matches the Reference in Encoding, Interpretation (ToM), and Goal. The reasoning logically justifies the Action.\\
 Rule: 1.0 for perfect alignment with the Reference's logic; 0.8-0.9 for correct logic with minor redundancy.\\
 Output Format:score (e.g., 0.7) \par

} \\

\bottomrule
\end{tabular}
\end{table*}

\label{app:2contentreward:data}

\paragraph{Teacher references (gold SIP rationales).}
We first provide the teacher model (o3) with the gold final answer and the original narrative context,
and obtain stage-consistent SIP rationales as silver references for each intermediate SIP stage.

\paragraph{Trajectory sampling from diverse checkpoints.}
To ensure a wide spectrum of reasoning quality, we collect candidate trajectories from 10 checkpoints of the
\textit{w/o $R_{\text{struct}}$} run at steps
$\{30, 90, 120, 180, 270, 360, 420, 510, 570, 600\}$.
We sample $K{=}6$ trajectories per instance per checkpoint.
Given $N{=}700$ training instances, this yields: 42000 candidate SIP-stage reasoning segments, spanning from early unstable logic to mature social inference.

\paragraph{LLM-based scoring with teacher calibration.}
Each sampled segment is scored by GPT-5 using the o3 silver rationale as the reference.
The evaluation rubric focuses on:
(i) fact-grounding to story-internal evidence,
(ii) correctness of mental-state attribution, and
(iii) stage-specific relevance within the SIP trajectory.

\paragraph{Tiering and hard-negative identification.}
We categorise all segments into five quality tiers:
\textbf{Tier S (Expert)}: o3 gold-quality samples.
\textbf{Tier A (Quality Positive)}: $\mathrm{ACC}{=}1$ and $\mathrm{LLM\_Score} \ge 0.8$.
\textbf{Tier B (Weak Positive)}: $\mathrm{ACC}{=}1$ and  $ \ge 0.6 \mathrm{LLM\_Score} \le 0.8$.
\textbf{Tier C (Hard Negative)}: $\mathrm{ACC}{=}1$ while $\mathrm{LLM\_Score} \le 0.6$  .
\textbf{Tier D (Common Negative)}: $\mathrm{ACC}{=}0$ and $\mathrm{LLM\_Score}$ is very low.

We construct preference pairs (\textit{chosen} vs.\ \textit{rejected}) using the following priority scheme:
 \textbf{P0}: Tier S $>$ Tier C \quad (strongest supervision against hard negatives).
 \textbf{P1}: Tier A $>$ Tier C.
  \textbf{P2}: Tier A $>$ Tier B.
   \textbf{P3}: Tier B $>$ Tier D.
  \textbf{P4}: late-stage short trajectories $>$ early-stage long trajectories,
        encouraging concise and cognitively disciplined inference.

Applying this policy produces \textbf{SocialPairs-20K}, consisting of 20K preference pairs
for training $RM_{\text{content}}$.

\paragraph{Held-out evaluation.}
We construct a held-out test set of  2K preference pairs from ToMBench-Hard test instances
using the same pipeline, achieving 89.2\% pairwise accuracy.
Additionally, we randomly sample 200 pairs for expert human verification,
obtaining 87.5\% agreement with human preferences.

The detailed multi-dimensional scoring rubric used by GPT-5 for stage-wise segment evaluation is provided in Table~\ref{tab:content_rubric}.

\section{Implementation }

\label{app:3implementation}
The content reward model (TRM) is initialized from Qwen3-4B-Instruct-2507and trained using supervised fine-tuning for two epochs on four NVIDIA A100 80GB GPUs. The model is trained with pairwise comparisons and, during inference, predicts a scalar reward conditioned on (\emph{scenario, question, options, reasoning}).

We evaluate multiple LLM families, including  
DeepSeek-R1\citep{guo2025deepseek}, 
DeepSeek-R1-Distill-Llama-70B\citep{guo2025deepseek}\footnote{\href{https://huggingface.co/deepseek-ai/DeepSeek-R1-Distill-Llama-70B}{https://huggingface.co/DeepSeek-R1-DistillLlama-70B}},
Llama-3.1-70B-Instruct\citep{dubey2024llama}\footnote{\href{https://huggingface.co/collections/meta-llama/llama-31-669fc079a0c406a149a5738f}{https://huggingface.co/collections/meta-llama-3-1}},  
Qwen3-4B/8B/32B\citep{yang2025qwen3}\footnote{\href{https://huggingface.co/collections/Qwen/qwen3-67dd247413f0e2e4f653967f}{https://huggingface.co/collections/Qwen/qwen3}}, 
GPT-5-2025-08-07\citep{ouyang2022training}, GPT-4o-2024-08-06 \citep{achiam2023gpt}, and o3-2025-04-16. We prompt Qwen3-4B/8B/32B in both Thinking and  No-Think setting.

\begin{figure}[h]
	\centering
	\setlength{\abovecaptionskip}{1mm}
	\setlength{\belowcaptionskip}{1mm}
	\includegraphics[width=1\linewidth]{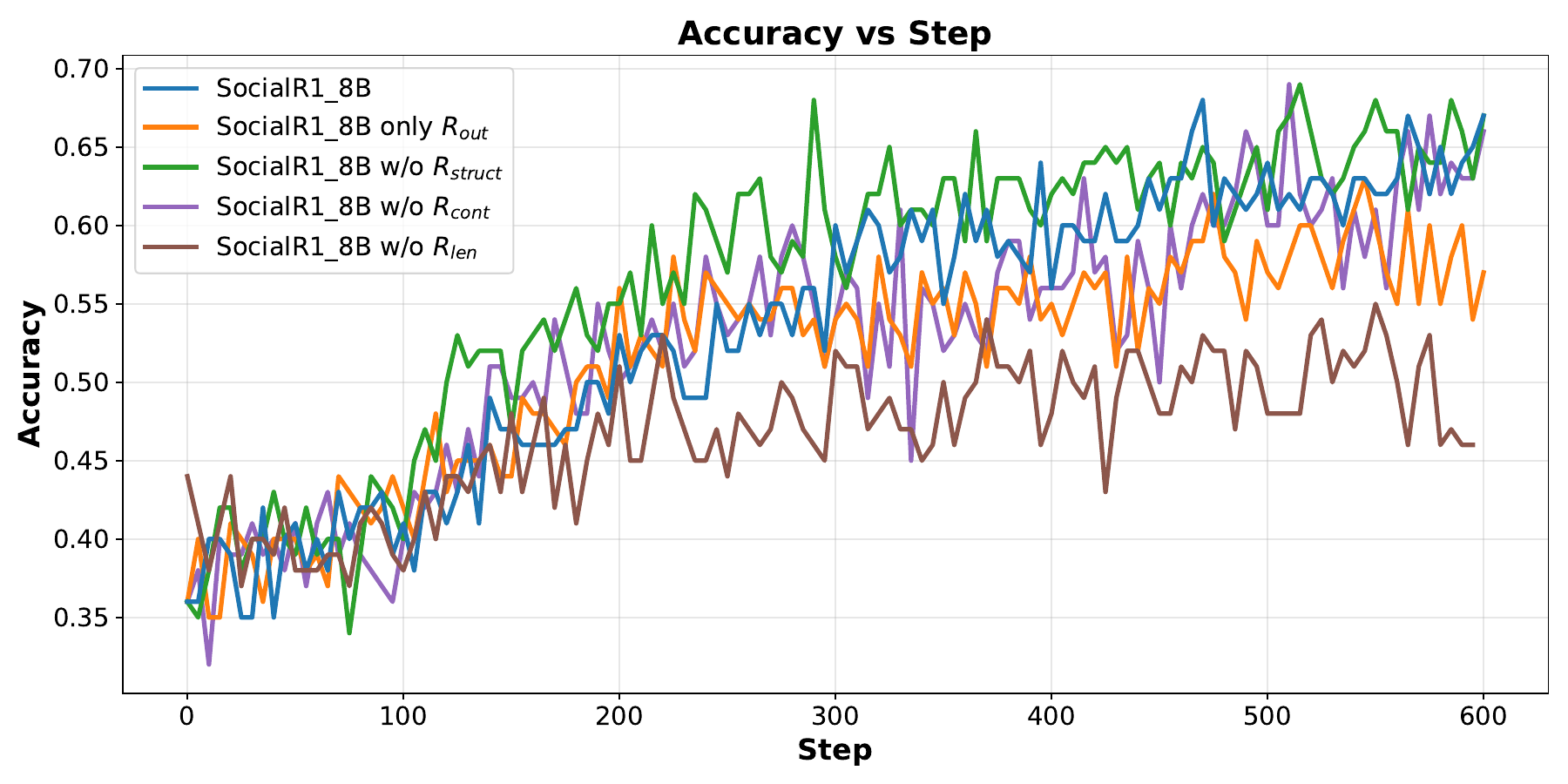}
	\caption{Training accuracy during training.}
	\label{fig-trainacc}
\end{figure}

\begin{figure}[h]
	\centering
	\setlength{\abovecaptionskip}{1mm}
	\setlength{\belowcaptionskip}{1mm}
	\includegraphics[width=1\linewidth]{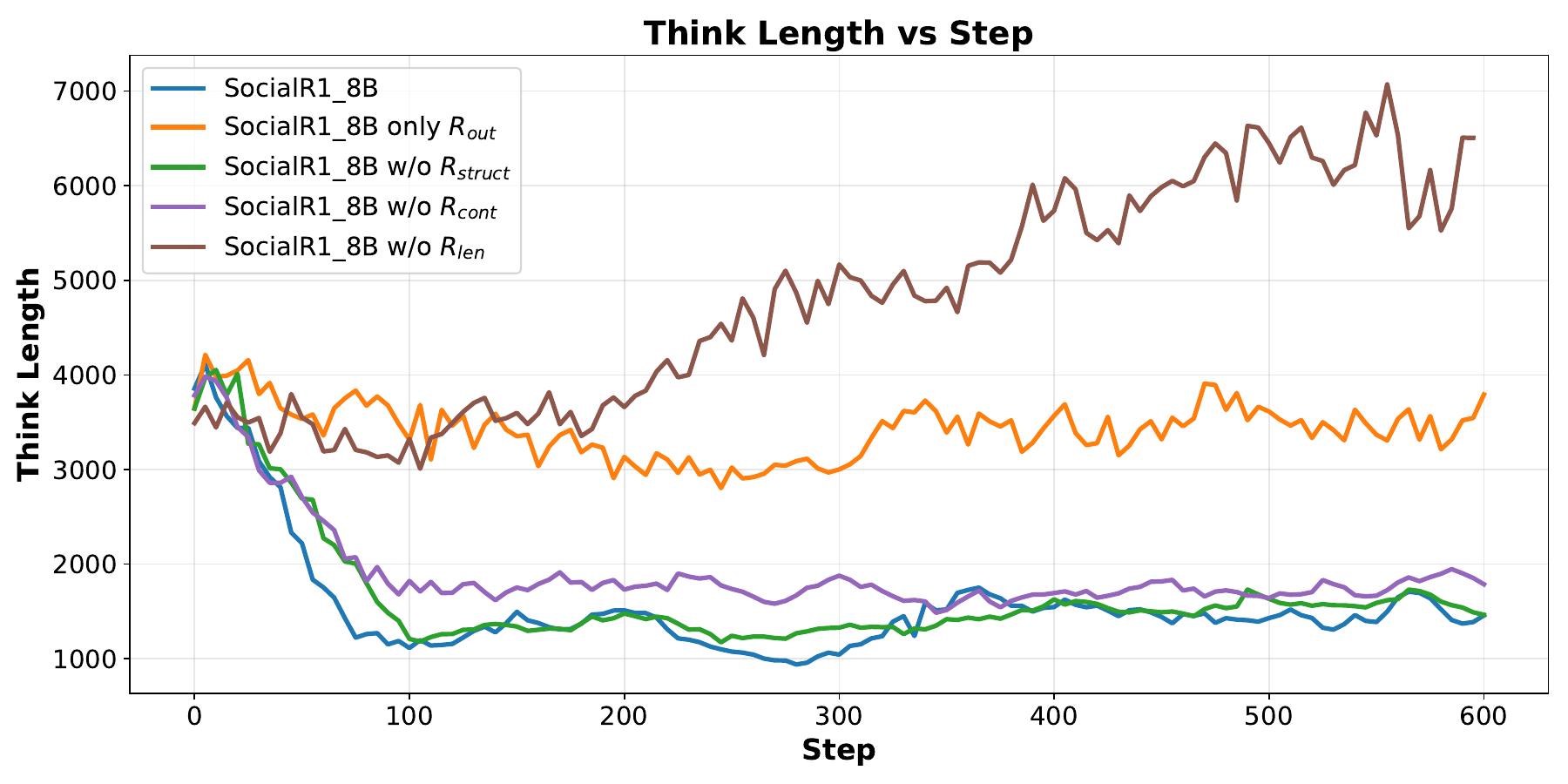}
	\caption{Trajectory length during training}
	\label{fig-trainlen}
\end{figure}

For completeness, we provide additional training dynamics in Appendix, including the evolution of training accuracy (Figure~\ref{fig-trainacc}), trajectory length (Figure~\ref{fig-trainlen}), content integrity score (Figure~\ref{fig-traincontent}), and structural alignment score during optimisation.

\begin{figure}[h]
	\centering
	\setlength{\abovecaptionskip}{1mm}
	\setlength{\belowcaptionskip}{1mm}
	\includegraphics[width=1\linewidth]{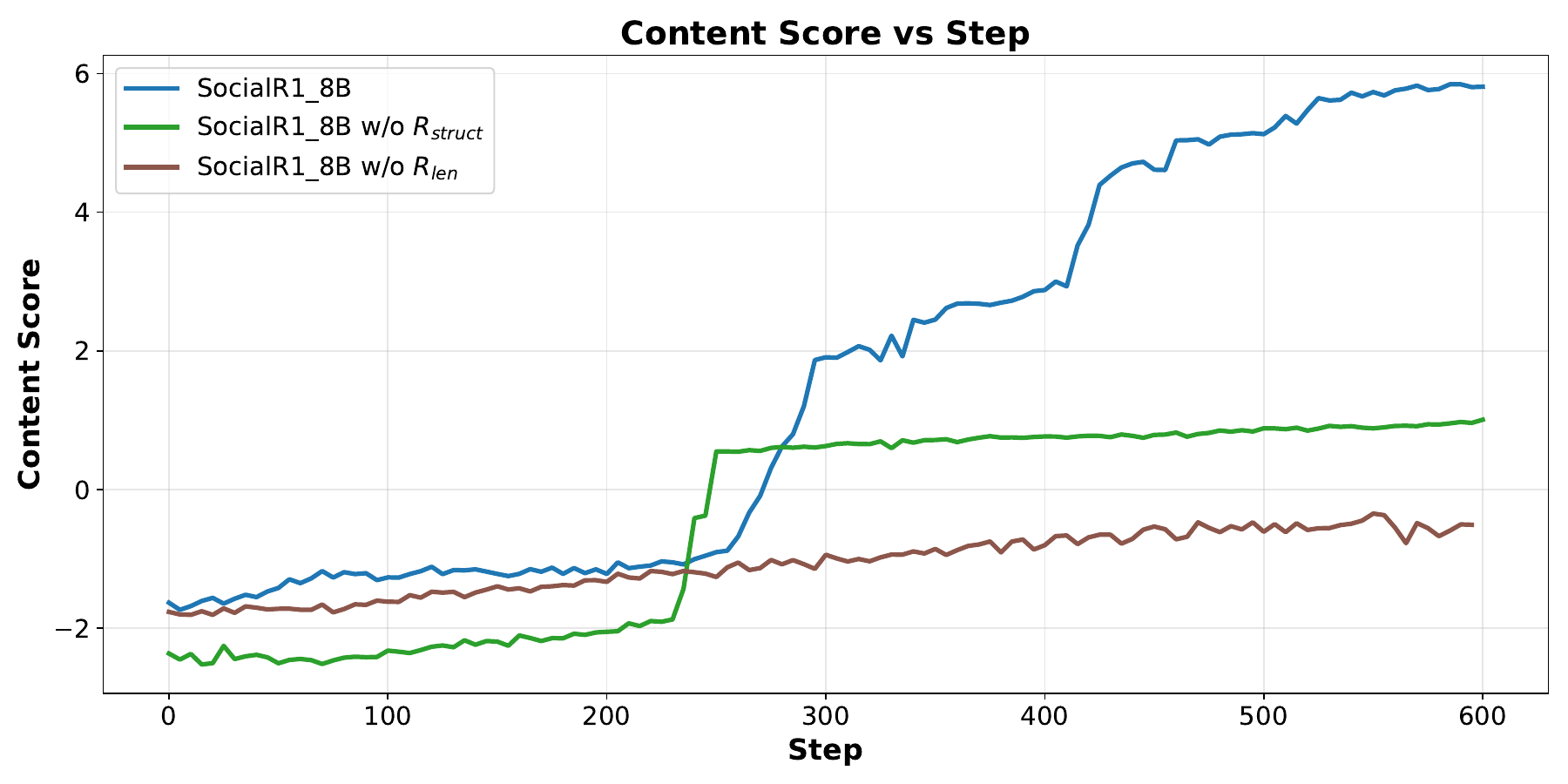}
	\caption{Content Score During Training}
	\label{fig-traincontent}
\end{figure}

\begin{figure}[h]
	\centering
	\setlength{\abovecaptionskip}{1mm}
	\setlength{\belowcaptionskip}{1mm}
	\includegraphics[width=1\linewidth]{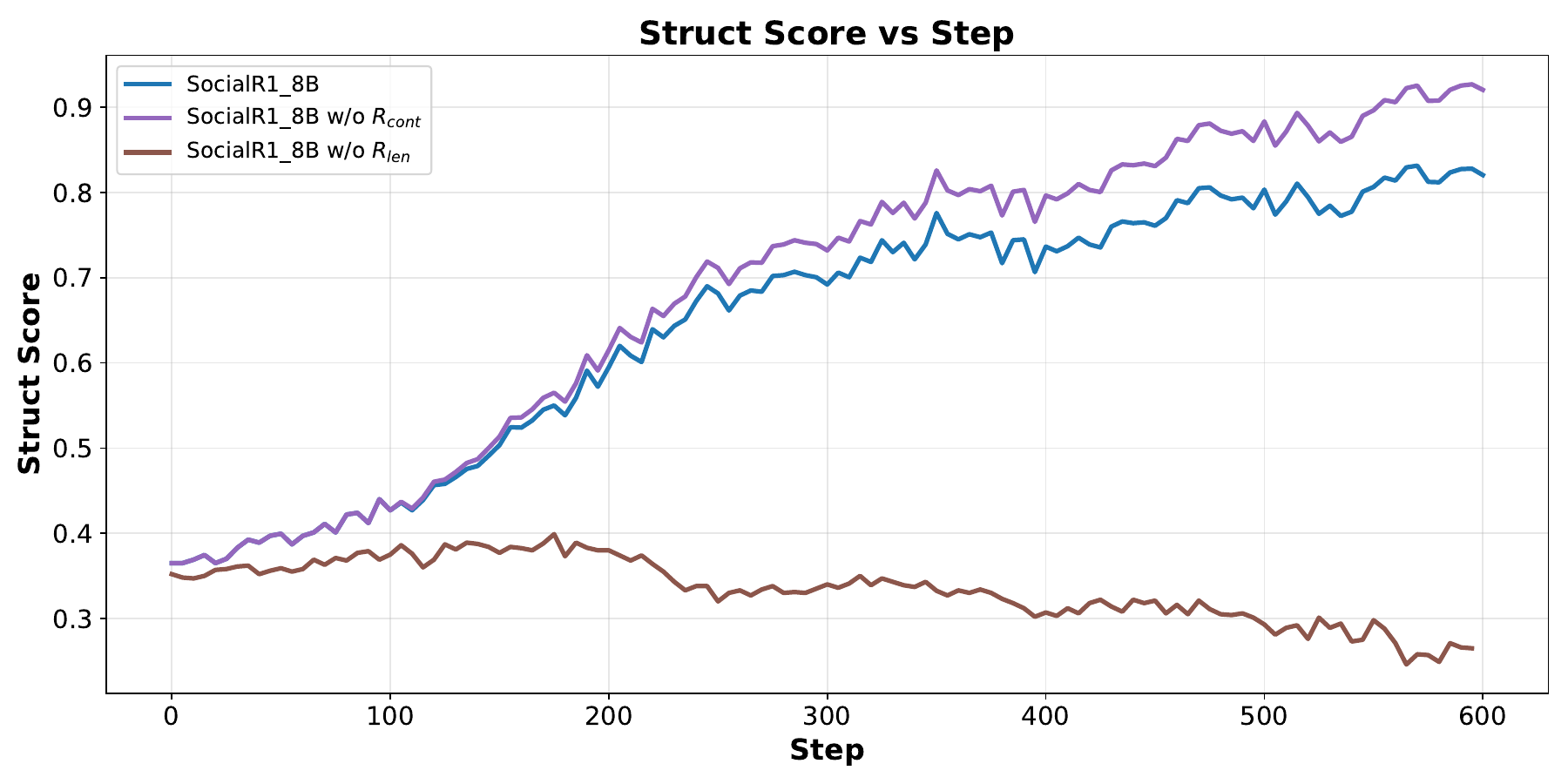}
	\caption{Structure Score During Training}
	\label{fig-traincontent}
\end{figure}


\section{Detailed Results}
\label{app:4mainresults}

For completeness, we report the full detailed results across all evaluated benchmarks in Appendix Tables~\ref{tab:motivbench_results}--\ref{tab:emotionbench_results}, 
including MotivationBench (Table~\ref{tab:motivbench_results}), 
ToMBench-Hard (Table~\ref{tab:tombench_hard_results}), 
SimpleToM (Table~\ref{tab:simpletom_results}), 
TactfulToM (Table~\ref{tab:tactfultom_results}), 
and EmoBench (Table~\ref{tab:emotionbench_results}).

\begin{table}[h]
\centering
\caption{Performance on MotiveBench}
\label{tab:motivbench_results}

\footnotesize
\setlength{\tabcolsep}{3.5pt}
\renewcommand{\arraystretch}{0.92}

\begin{tabular*}{\linewidth}{@{\extracolsep{\fill}}lcccc@{}}
\toprule
\textbf{Model} & \textbf{Amazon} & \textbf{Blog} & \textbf{Persona} & \textbf{Overall} \\
\midrule

\multicolumn{5}{l}{\textbf{Closed-sourced LLMs}}\\
\midrule
DeepSeek-R1   & 0.9000 & 0.8333 & 0.8633 & 0.8655 \\
o3            & 0.9800 & 0.9067 & 0.9333 & 0.9400 \\
o3\_COT       & 0.9667 & 0.8933 & 0.9000 & 0.9200 \\
GPT-5         & 0.9200 & 0.9133 & 0.8933 & 0.9089 \\
GPT-5\_COT    & 0.9600 & 0.9267 & 0.8933 & 0.9267 \\
GPT-4o        & 0.9733 & 0.9033 & 0.9133 & 0.9300 \\
GPT-4o\_COT   & 0.9400 & 0.8867 & 0.8667 & 0.8978 \\

\midrule
\multicolumn{5}{l}{\textbf{Open-sourced LLMs}}\\
\midrule
Qwen3-4B (Disable)   & 0.8067 & 0.7867 & 0.7267 & 0.7734 \\
Qwen3-4B             & 0.9133 & 0.7933 & 0.8267 & 0.8444 \\
Qwen3-8B (Disable)   & 0.8200 & 0.8000 & 0.7600 & 0.7933 \\
Qwen3-8B             & 0.7333 & 0.6600 & 0.6700 & 0.6878 \\
Qwen3-32B (Disable)  & 0.9333 & 0.8900 & 0.8533 & 0.8922 \\
Qwen3-32B            & 0.9067 & 0.8867 & 0.8467 & 0.8800 \\
LLaMa3.1-70B         & 0.9000 & 0.7867 & 0.8467 & 0.8445 \\
LLaMa3.1-70B\_COT    & 0.8933 & 0.8800 & 0.8867 & 0.8867 \\
Distill-LLaMa-70B    & 0.9333 & 0.8333 & 0.8867 & 0.8844 \\

\midrule
\multicolumn{5}{l}{\textbf{Ours}}\\
\midrule
SocialR1-4B only \Rout     & 0.8933 & 0.7933 & 0.8733 & 0.8533 \\
SocialR1-4B w/o \Rlen      & 0.8800 & 0.8133 & 0.8000 & 0.8311 \\
SocialR1-4B w/o \Rstruct   & 0.8667 & 0.7867 & 0.8400 & 0.8311 \\
SocialR1-4B w/o \Rcont     & 0.8667 & 0.7933 & 0.8300 & 0.8300 \\
SocialR1-4B Full           & 0.8733 & 0.8433 & 0.8333 & 0.8500 \\

\addlinespace[0.3ex]

SocialR1-8B only \Rout     & 0.9333 & 0.8667 & 0.8667 & 0.8889 \\
SocialR1-8B w/o \Rlen      & 0.9267 & 0.8333 & 0.8700 & 0.8767 \\
SocialR1-8B w/o \Rstruct   & 0.9000 & 0.8600 & 0.8767 & 0.8789 \\
SocialR1-8B w/o \Rcont     & 0.9000 & 0.8267 & 0.8433 & 0.8567 \\
SocialR1-8B Full           & 0.9000 & 0.8600 & 0.8667 & 0.8756 \\

\bottomrule
\end{tabular*}

\end{table}

\begin{table}[h]
\centering
\caption{Performance on SimpleToM.}
\label{tab:simpletom_results}

\small
\setlength{\tabcolsep}{4pt}
\renewcommand{\arraystretch}{1.0}

\resizebox{0.88\columnwidth}{!}{%
\begin{tabular}{lcccc}
\specialrule{1.2pt}{0pt}{0pt}

\textbf{Model} & \textbf{Behav} & \textbf{Judge} & \textbf{Mental} & \textbf{Overall} \\

\midrule
\multicolumn{5}{l}{\textbf{Closed-sourced LLMs}}\\
\midrule

DeepSeek-R1  & 0.6818 & 0.6452 & 0.8291 & 0.7187 \\
o3           & 0.8030 & 0.4996 & 0.9895 & 0.7640 \\
o3\_COT      & 0.8134 & 0.4560 & 0.9808 & 0.7501 \\
gpt5         & 0.7350 & 0.4908 & 0.9808 & 0.7355 \\
gpt5\_COT    & 0.6949 & 0.4551 & 0.9834 & 0.7111 \\
gpt5\_MS     & 0.9913 & 0.9887 & 0.9974 & 0.9924 \\
gpt4o        & 0.5257 & 0.6957 & 0.7768 & 0.6661 \\
gpt4o\_COT   & 0.5684 & 0.3339 & 0.9355 & 0.6126 \\
gpt4o\_MS    & 0.8326 & 0.4010 & 0.9738 & 0.7358 \\

\midrule
\multicolumn{5}{l}{\textbf{Open-sourced LLMs}}\\
\midrule

Qwen3-4B (Disable)   & 0.4673 & 0.2415 & 0.8239 & 0.5109 \\
Qwen3-4B             & 0.4987 & 0.2214 & 0.9032 & 0.5411 \\
Qwen3-8B (Disable)   & 0.4673 & 0.2415 & 0.8239 & 0.5109 \\
Qwen3-8B             & 0.5711 & 0.3374 & 0.9442 & 0.6176 \\
Qwen3-32B (Disable)  & 0.5719 & 0.5484 & 0.8230 & 0.6478 \\
Qwen3-32B            & 0.6591 & 0.6678 & 0.9634 & 0.7634 \\
LLaMa3.1-70B         & 0.6722 & 0.5196 & 0.9651 & 0.7190 \\
LLaMa3.1-70B\_COT    & 0.5371 & 0.3810 & 0.8544 & 0.5908 \\
Distill-LLaMa-70B    & 0.7280 & 0.7358 & 0.8248 & 0.7629 \\

\midrule
\multicolumn{5}{l}{\textbf{Ours}}\\
\midrule

SocialR1-4B only \Rout        & 0.8535 & 0.9085 & 0.9268 & 0.8963 \\
SocialR1-4B w/o \Rlen       & 0.8692 & 0.9268 & 0.9634 & 0.9198 \\
SocialR1-4B w/o \Rstruct    & 0.8910 & 0.9948 & 0.9625 & 0.9494 \\
SocialR1-4B w/o \Rcont      & 0.7716 & 0.9991 & 0.9172 & 0.8960 \\
SocialR1-4B Full            & 0.9207 & 1.0000 & 0.8649 & 0.9285 \\

\midrule

SocialR1-8B only \Rout        & 0.8649 & 0.9861 & 0.9791 & 0.9433 \\
SocialR1-8B w/o \Rlen       & 0.9163 & 0.9948 & 0.9747 & 0.9619 \\
SocialR1-8B w/o \Rstruct    & 0.8265 & 0.9930 & 0.8928 & 0.9041 \\
SocialR1-8B w/o \Rcont      & 0.8361 & 0.9826 & 0.9869 & 0.9352 \\
SocialR1-8B Full            & 0.9198 & 0.9948 & 0.9878 & 0.9675 \\

\specialrule{1.2pt}{0pt}{0pt}
\end{tabular}
}
\end{table}

\begin{table*}[t]
\centering
\caption{Performances on ToMBench}
\label{tab:tombench_full}

\footnotesize
\setlength{\tabcolsep}{3pt}
\renewcommand{\arraystretch}{0.88}

\resizebox{\textwidth}{!}{%
\begin{tabular}{lccccccc}
\specialrule{1.2pt}{0pt}{0pt}

\textbf{Model} &
\textbf{Belief} &
\textbf{Desire} &
\textbf{Emotion} &
\textbf{Intention} &
\textbf{Knowledge} &
\textbf{\makecell{Non-literal\\Comm.}} &
\textbf{Overall} \\

\midrule
\multicolumn{8}{l}{\textbf{Closed-sourced LLMs}}\\
\midrule

DeepSeek-R1   & 0.7846 & 0.5611 & 0.6905 & 0.7824 & 0.6332 & 0.6484 & 0.6834 \\
o3            & 0.9252 & 0.7111 & 0.7786 & 0.8529 & 0.6263 & 0.8062 & 0.7834 \\
o3\_COT     & 0.8821 & 0.6500 & 0.7548 & 0.8147 & 0.5952 & 0.7447 & 0.7402 \\
GPT5.2        & 0.9274 & 0.6611 & 0.7952 & 0.8706 & 0.6332 & 0.8062 & 0.7823 \\
GPT5.2\_COT    & 0.9151 & 0.6833 & 0.7929 & 0.8618 & 0.6401 & 0.8048 & 0.7830 \\
GPT4          & 0.8594 & 0.6278 & 0.7524 & 0.8235 & 0.5779 & 0.7861 & 0.7378 \\
GPT4\_COT      & 0.8980 & 0.6278 & 0.7405 & 0.7971 & 0.5675 & 0.8128 & 0.7406 \\

\midrule
\multicolumn{8}{l}{\textbf{Open-sourced LLMs}}\\
\midrule

Qwen3-4B (Disable thinking)   & 0.6168 & 0.5444 & 0.6262 & 0.6324 & 0.3322 & 0.7099 & 0.5770 \\
Qwen3-4B             & 0.7506 & 0.5722 & 0.6810 & 0.6882 & 0.3945 & 0.7273 & 0.6356 \\
Qwen3-8B (Disable thinking)   & 0.6383 & 0.5000 & 0.6262 & 0.6441 & 0.2595 & 0.5414 & 0.5349 \\
Qwen3-8B             & 0.7982 & 0.5889 & 0.6667 & 0.7176 & 0.2664 & 0.6698 & 0.6179 \\
Qwen3-32B (Disable thinking)  & 0.8073 & 0.6167 & 0.7167 & 0.8147 & 0.4394 & 0.7553 & 0.6917 \\
Qwen3-32B                     & 0.8560 & 0.6111 & 0.7286 & 0.8324 & 0.4014 & 0.7553 & 0.6975 \\
LLaMa3.1-70B                  & 0.7868 & 0.5000 & 0.6286 & 0.6529 & 0.3875 & 0.6658 & 0.6036 \\
LLaMa3.1-70B\_COT           & 0.8118 & 0.5222 & 0.7357 & 0.8324 & 0.5467 & 0.7246 & 0.6956 \\
Distill-LLaMa-70B             & 0.8798 & 0.6389 & 0.7429 & 0.8441 & 0.5675 & 0.7660 & 0.7399 \\

\midrule
\multicolumn{8}{l}{\textbf{Ours}}\\
\midrule

SocialR1-4B only  \Rout          & 0.5612 & 0.5333 & 0.6286 & 0.6324 & 0.5225 & 0.6791 & 0.5929 \\
SocialR1-4B w/o \Rlen  & 0.7914 & 0.5500 & 0.6000 & 0.6941 & 0.4671 & 0.7233 & 0.6376 \\
SocialR1-4B w/o \Rstruct  & 0.6349 & 0.5778 & 0.6524 & 0.6676 & 0.6055 & 0.7313 & 0.6449 \\
SocialR1-4B w/o \Rcont & 0.6327 & 0.5833 & 0.6190 & 0.7353 & 0.5571 & 0.7420 & 0.6449 \\
SocialR1-4B Full   & 0.7358 & 0.5444 & 0.6929 & 0.6853 & 0.5882 & 0.6979 & 0.6574 \\

\midrule

SocialR1-8B only  \Rout       & 0.8243 & 0.6056 & 0.6881 & 0.7500 & 0.5190 & 0.7233 & 0.6850 \\
SocialR1-8B w/o \Rlen       & 0.7517 & 0.5611 & 0.6667 & 0.7912 & 0.5882 & 0.7580 & 0.6862 \\
SocialR1-8B w/o \Rstruct    & 0.6349 & 0.5778 & 0.6524 & 0.6676 & 0.6055 & 0.7313 & 0.6449 \\
SocialR1-8B w/o \Rcont      & 0.7517 & 0.5611 & 0.6762 & 0.7588 & 0.6125 & 0.7487 & 0.6848 \\
SocialR1-8B Full        & 0.7778 & 0.5444 & 0.6524 & 0.8462 & 0.5848 & 0.7233 & 0.6881 \\

\specialrule{1.2pt}{0pt}{0pt}
\end{tabular}
}
\end{table*}

\begin{table*}[h]
\centering
\caption{Performance on TactfulToM.}
\label{tab:tactfultom_results}

\scriptsize
\setlength{\tabcolsep}{2pt}
\renewcommand{\arraystretch}{0.78}

\resizebox{\textwidth}{!}{%
\begin{tabular}{p{2.7cm}ccccccccc}
\specialrule{1.1pt}{0pt}{0pt}

\textbf{Model} &
\textbf{Ans.} &
\textbf{Belief} &
\textbf{Comp.} &
\textbf{Fact-R} &
\textbf{Fact-T} &
\textbf{\makecell{Info\\Acc.}} &
\textbf{Liab.} &
\textbf{Detect.} &
\textbf{Overall} \\

\midrule
\multicolumn{10}{l}{\textbf{Closed-sourced LLMs}}\\
\midrule

DeepSeek-R1  & 0.4036 & 0.7629 & 0.7300 & 0.9300 & 0.8529 & 0.6413 & 0.8333 & 0.4214 & 0.6969 \\
o3           & 0.3353 & 0.7951 & 0.7100 & 0.9600 & 0.8529 & 0.7803 & 0.7273 & 0.3267 & 0.6860 \\
o3\_COT      & 0.3095 & 0.7599 & 0.6800 & 0.8700 & 0.8088 & 0.7713 & 0.6919 & 0.3433 & 0.6544 \\
GPT5         & 0.3254 & 0.7138 & 0.2800 & 0.9800 & 0.7942 & 0.4686 & 0.6313 & 0.3367 & 0.4670 \\
GPT5\_COT    & 0.3770 & 0.7569 & 0.6800 & 0.9700 & 0.7794 & 0.6143 & 0.6869 & 0.3733 & 0.6547 \\
GPT4         & 0.3254 & 0.5938 & 0.2700 & 0.9900 & 0.6912 & 0.4395 & 0.5707 & 0.3333 & 0.5267 \\
GPT4\_COT    & 0.3790 & 0.6245 & 0.3400 & 0.9400 & 0.5588 & 0.5426 & 0.4798 & 0.3600 & 0.5281 \\

\midrule
\multicolumn{10}{l}{\textbf{Open-sourced LLMs}}\\
\midrule

Qwen3-4B (Disable)   & 0.3571 & 0.3512 & 0.1800 & 0.8000 & 0.4559 & 0.4327 & 0.2778 & 0.3700 & 0.4031 \\
Qwen3-4B             & 0.4147 & 0.4583 & 0.2000 & 0.9000 & 0.5000 & 0.4260 & 0.3283 & 0.4400 & 0.4584 \\
Qwen3-8B (Disable)   & 0.3333 & 0.3914 & 0.2000 & 0.9100 & 0.4559 & 0.4103 & 0.3990 & 0.3333 & 0.4292 \\
Qwen3-8B             & 0.3988 & 0.4578 & 0.2900 & 0.9600 & 0.4118 & 0.4265 & 0.4394 & 0.3400 & 0.4655 \\
Qwen3-32B (Disable)  & 0.3750 & 0.5035 & 0.3500 & 0.9100 & 0.6324 & 0.4686 & 0.5152 & 0.4000 & 0.5193 \\
Qwen3-32B            & 0.3571 & 0.6042 & 0.4400 & 0.9600 & 0.6618 & 0.5516 & 0.5354 & 0.3500 & 0.5575 \\
LLaMa3.1-70B         & 0.3750 & 0.5461 & 0.5000 & 0.9900 & 0.5294 & 0.5336 & 0.5707 & 0.3633 & 0.5510 \\
LLaMa3.1-70B\_COT    & 0.3730 & 0.4722 & 0.8600 & 0.9000 & 0.5000 & 0.6614 & 0.3232 & 0.3767 & 0.5583 \\
Distill-LLaMa-70B    & 0.5734 & 0.6136 & 0.6100 & 0.8700 & 0.6912 & 0.6188 & 0.4646 & 0.5200 & 0.6202 \\

\midrule
\multicolumn{10}{l}{\textbf{Ours}}\\
\midrule

SocialR1-4B wo \Rout        & 0.3571 & 0.3943 & 0.2600 & 0.8900 & 0.4853 & 0.4395 & 0.2121 & 0.3733 & 0.4265 \\
SocialR1-4B w/o \Rlen       & 0.4246 & 0.4415 & 0.2600 & 0.9300 & 0.5735 & 0.4170 & 0.4596 & 0.4467 & 0.4941 \\
SocialR1-4B w/o \Rstruct    & 0.4147 & 0.5124 & 0.1100 & 0.9500 & 0.6176 & 0.4910 & 0.4949 & 0.4100 & 0.5001 \\
SocialR1-4B w/o \Rcont      & 0.3552 & 0.3993 & 0.3500 & 0.9100 & 0.4118 & 0.4350 & 0.4596 & 0.3667 & 0.4609 \\
SocialR1-4B Full            & 0.4881 & 0.4296 & 0.2600 & 0.9600 & 0.4559 & 0.4260 & 0.4596 & 0.4700 & 0.4936 \\

\midrule
SocialR1-8B wo \Rout        & 0.3512 & 0.4871 & 0.2900 & 0.9700 & 0.5000 & 0.4955 & 0.4040 & 0.3900 & 0.4860 \\
SocialR1-8B w/o \Rlen       & 0.3790 & 0.4732 & 0.2300 & 0.9600 & 0.5147 & 0.5650 & 0.4091 & 0.3667 & 0.4872 \\
SocialR1-8B w/o \Rstruct    & 0.3433 & 0.3403 & 0.3000 & 0.9100 & 0.4559 & 0.5179 & 0.4293 & 0.3500 & 0.4558 \\
SocialR1-8B w/o \Rcont      & 0.3393 & 0.3363 & 0.3900 & 0.9300 & 0.4265 & 0.5673 & 0.4040 & 0.3400 & 0.4667 \\
SocialR1-8B Full            & 0.3512 & 0.4122 & 0.3900 & 0.9200 & 0.6176 & 0.5987 & 0.3838 & 0.3900 & 0.5079 \\

\specialrule{1.1pt}{0pt}{0pt}
\end{tabular}
}
\end{table*}


\begin{table*}[t]
\centering
\caption{Performances on EmoBench}
\label{tab:emotionbench_results}

\scriptsize
\setlength{\tabcolsep}{2pt}
\renewcommand{\arraystretch}{0.78}

\resizebox{\textwidth}{!}{%
\begin{tabular}{lccccccc}
\specialrule{1.1pt}{0pt}{0pt}

\textbf{Model} &
\textbf{\makecell{Complex\\Emotions}} &
\textbf{\makecell{Emotional\\Cues}} &
\textbf{\makecell{Personal Beliefs\\\& Experiences}} &
\textbf{\makecell{Perspective\\Taking}} &
\textbf{Interpersonal} &
\textbf{Self} &
\textbf{Overall} \\

\midrule
\multicolumn{8}{l}{\textbf{Closed-sourced LLMs}}\\
\midrule

DeepSeek-R1  & 0.7959 & 0.7857 & 0.7143 & 0.7164 & 0.7000 & 0.7200 & 0.7387 \\
o3           & 0.8571 & 0.8393 & 0.8036 & 0.8134 & 0.7600 & 0.7500 & 0.8039 \\
o3\_COT      & 0.7653 & 0.7500 & 0.7232 & 0.7164 & 0.6900 & 0.6900 & 0.7225 \\
GPT5         & 0.8673 & 0.8571 & 0.7946 & 0.7687 & 0.7800 & 0.7500 & 0.8030 \\
GPT5\_COT    & 0.8571 & 0.7679 & 0.7679 & 0.7761 & 0.7800 & 0.7800 & 0.7882 \\
GPT4         & 0.8469 & 0.7857 & 0.7321 & 0.7015 & 0.7800 & 0.7700 & 0.7694 \\
GPT4\_COT    & 0.8163 & 0.8214 & 0.7411 & 0.7687 & 0.7800 & 0.7300 & 0.7762 \\

\midrule
\multicolumn{8}{l}{\textbf{Open-sourced LLMs}}\\
\midrule

Qwen3-4B (Disable)      & 0.5306 & 0.5179 & 0.4643 & 0.4254 & 0.5800 & 0.6500 & 0.5280 \\
Qwen3-4B                & 0.5612 & 0.5714 & 0.4821 & 0.4403 & 0.5500 & 0.5800 & 0.5308 \\
Qwen3-8B (Disable)      & 0.7041 & 0.7500 & 0.5893 & 0.5373 & 0.5700 & 0.7400 & 0.6484 \\
Qwen3-8B                & 0.5714 & 0.6607 & 0.4821 & 0.4179 & 0.7600 & 0.6600 & 0.5920 \\
Qwen3-32B (Disable)     & 0.6939 & 0.7321 & 0.5893 & 0.5448 & 0.6800 & 0.7800 & 0.6700 \\
Qwen3-32B               & 0.6531 & 0.6429 & 0.5982 & 0.5597 & 0.6600 & 0.7000 & 0.6356 \\
LLaMa3.1-70B            & 0.7143 & 0.8214 & 0.7143 & 0.6194 & 0.5000 & 0.6100 & 0.6632 \\
LLaMa3.1-70B\_COT       & 0.6837 & 0.8036 & 0.6607 & 0.6567 & 0.6500 & 0.6800 & 0.6891 \\
Distill-LLaMa-70B       & 0.7959 & 0.8214 & 0.7411 & 0.7090 & 0.7600 & 0.7300 & 0.7596 \\

\midrule
\multicolumn{8}{l}{\textbf{Ours}}\\
\midrule

SocialR1-4B only  \Rout\        & 0.6735 & 0.6429 & 0.6250 & 0.5821 & 0.6400 & 0.6900 & 0.6422 \\
SocialR1-4B w/o  \Rlen\      & 0.7041 & 0.6607 & 0.6875 & 0.5522 & 0.6200 & 0.7300 & 0.6591 \\
SocialR1-4B w/o  \Rstruct\      & 0.6939 & 0.7857 & 0.6250 & 0.5896 & 0.6900 & 0.6800 & 0.6774 \\
SocialR1-4B w/o \Rcont\    & 0.6633 & 0.6429 & 0.6250 & 0.5597 & 0.5800 & 0.6800 & 0.6251 \\
SocialR1-4B Full     & 0.7041 & 0.7500 & 0.7143 & 0.6642 & 0.5800 & 0.6700 & 0.6804 \\

\midrule

SocialR1-8B only  \Rout\        & 0.7449 & 0.7321 & 0.7143 & 0.6194 & 0.6500 & 0.7200 & 0.6968 \\
SocialR1-8B w/o \Rlen\     & 0.7653 & 0.8036 & 0.6875 & 0.6194 & 0.5400 & 0.7000 & 0.6860 \\
SocialR1-8B w/o \Rstruct\      & 0.7551 & 0.7321 & 0.6607 & 0.6418 & 0.6200 & 0.7900 & 0.7000 \\
SocialR1-8B w/o \Rcont\    & 0.7245 & 0.7857 & 0.6696 & 0.6269 & 0.5200 & 0.7400 & 0.6778 \\
SocialR1-8B  Full   & 0.7449 & 0.7857 & 0.6964 & 0.6493 & 0.6100 & 0.7200 & 0.7010 \\

\specialrule{1.1pt}{0pt}{0pt}
\end{tabular}
}
\end{table*}

\begin{table*}[h]
\centering
\caption{Performance on ToMBench\_Hard Validation Set}
\label{tab:tombench_hard_results}

\scriptsize
\setlength{\tabcolsep}{2pt}
\renewcommand{\arraystretch}{0.78}

\resizebox{\textwidth}{!}{%
\begin{tabular}{lccccccc}
\specialrule{1.1pt}{0pt}{0pt}

\textbf{Model} &
\textbf{Belief} &
\textbf{Desire} &
\textbf{Emotion} &
\textbf{Intention} &
\textbf{Knowledge} &
\textbf{\makecell{Non-literal\\Comm.}} &
\textbf{Overall} \\

\midrule
\multicolumn{8}{l}{\textbf{Closed-sourced LLMs}}\\
\midrule

DeepSeek-R1  & 0.5000 & 0.4167 & 0.8500 & 0.3590 & 0.2857 & 0.2500 & 0.4436 \\
o3           & 0.5000 & 0.3333 & 0.6500 & 0.4359 & 0.1429 & 0.6250 & 0.4478 \\
o3\_COT      & 0.5714 & 0.5000 & 0.6500 & 0.3846 & 0.2857 & 0.5000 & 0.4820 \\
GPT5         & 0.2857 & 0.2500 & 0.6500 & 0.3590 & 0.2857 & 0.6250 & 0.4092 \\
GPT5\_COT    & 0.2857 & 0.3333 & 0.7000 & 0.3333 & 0.2857 & 0.6250 & 0.4272 \\
GPT4         & 0.2143 & 0.5000 & 0.7000 & 0.3077 & 0.1429 & 0.3750 & 0.3733 \\
GPT4\_COT    & 0.2857 & 0.2500 & 0.6500 & 0.4359 & 0.2857 & 0.5000 & 0.4012 \\

\midrule
\multicolumn{8}{l}{\textbf{Open-sourced LLMs}}\\
\midrule

Qwen3-4B (Disable)     & 0.1429 & 0.3333 & 0.5000 & 0.4359 & 0.0000 & 0.5000 & 0.3187 \\
Qwen3-4B               & 0.2143 & 0.2500 & 0.5500 & 0.3846 & 0.1429 & 0.5000 & 0.3403 \\
Qwen3-8B (Disable)     & 0.3571 & 0.3333 & 0.7500 & 0.3590 & 0.1429 & 0.3750 & 0.3862 \\
Qwen3-8B               & 0.1429 & 0.2500 & 0.6500 & 0.4359 & 0.2857 & 0.6250 & 0.3982 \\
Qwen3-32B (Disable)    & 0.1429 & 0.3333 & 0.7000 & 0.3846 & 0.2857 & 0.6250 & 0.4119 \\
Qwen3-32B              & 0.2143 & 0.3333 & 0.7500 & 0.4103 & 0.2857 & 0.6250 & 0.4364 \\
LLaMa3.1-70B           & 0.2857 & 0.4167 & 0.8500 & 0.4872 & 0.1429 & 0.5000 & 0.4471 \\
LLaMa3.1-70B\_COT      & 0.2143 & 0.3333 & 0.7500 & 0.4103 & 0.1429 & 0.7500 & 0.4335 \\
Distill-LLaMa-70B      & 0.2143 & 0.3333 & 0.7500 & 0.4615 & 0.2857 & 0.3750 & 0.4033 \\

\midrule
\multicolumn{8}{l}{\textbf{Ours}}\\
\midrule

SocialR1-4B only \Rout         & 0.3571 & 0.1667 & 0.6000 & 0.7436 & 0.0000 & 0.5000 & 0.3946 \\
SocialR1-4B w/o \Rlen    & 0.4286 & 0.2500 & 0.6500 & 0.6154 & 0.1429 & 0.7500 & 0.4728 \\
SocialR1-4B w/o \Rstruct   & 0.3571 & 0.1667 & 0.7000 & 0.7692 & 0.2857 & 0.3750 & 0.4423 \\
SocialR1-4B w/o \Rcont    & 0.7143 & 0.4167 & 0.6500 & 0.6923 & 0.2857 & 0.8750 & 0.4423 \\
SocialR1-4B Full     & 0.4286 & 0.2500 & 0.7000 & 0.7436 & 0.2857 & 0.5000 & 0.4846 \\

\midrule

SocialR1-8B only \Rout         & 0.1429 & 0.2500 & 0.7000 & 0.7179 & 0.7143 & 0.5000 & 0.5042 \\
SocialR1-8B w/o \Rlen    & 0.1429 & 0.3333 & 0.7500 & 0.7179 & 0.7143 & 0.5000 & 0.5264 \\
SocialR1-8B w/o \Rstruct    & 0.6429 & 0.3333 & 0.6000 & 0.7436 & 0.8571 & 0.5000 & 0.6128 \\
SocialR1-8B w/o \Rcont   & 0.3571 & 0.3333 & 0.8000 & 0.7436 & 0.5714 & 0.5000 & 0.5509 \\
SocialR1-8B Full     & 0.5000 & 0.3333 & 0.8000 & 0.7949 & 0.7143 & 0.6250 & 0.6279 \\

\specialrule{1.1pt}{0pt}{0pt}
\end{tabular}
}
\end{table*}

\section{Case Study}
\label{app:5optionmatching}

For completeness, we provide the full qualitative case study examples corresponding to Figure~\ref{fig:cases}. 
Tables~\ref{tab:case_socialR1_8B}--\ref{tab:case_deepseekr1_2} present the complete narrative contexts, questions, answer options, and model-generated SIP trajectories across Social-R1-8B and its reward-ablated variants. 
These detailed instances allow a closer inspection of the mechanistic failure modes identified in the main text, including ungrounded cue interpretation, goal misidentification, and option-level lexical shortcutting under outcome-only supervision. 
Together, these supplementary cases offer concrete evidence that Social-R1’s improvements arise from enforcing evidence-grounded interpretation and stage-consistent social reasoning, rather than answer-driven backfilling or superficial option matching.


\begin{table*}[h]
\centering
\caption{Case Study SocialR1-8B}
\label{tab:case_socialR1_8B}

\small
\setlength{\tabcolsep}{10pt}
\renewcommand{\arraystretch}{1.2}
\setlength{\emergencystretch}{2em} 

\begin{tabular}{p{0.92\textwidth}}
\toprule
\textbf{Case Study SocialR1-8B} \\
\midrule

{\raggedright
\textbf{[Story]} In the mysterious underground world of Terra Valley lives a small, mischievous robot named Grimmo.
In this realm without sky or celestial bodies, Grimmo has never seen the heavens, nor encountered a human.
Yet Terra Valley is a wonder in itself---its walls adorned with bioluminescent fungi and glimmering minerals that sparkle in the dark.
One day, Grimmo begins an imitation routine: extending his arms, he slowly spins on his own axis, his movements reminiscent of a spinning top
or a graceful dancer twirling in place.\par

\textbf{[Question]} What is Grimmo possibly imitating?\par

\textbf{[Option]} A. A floating cloud \quad
B. The rotation of a planet \quad
C. A dancer's spiral turn \quad
D. A glowing mushroom spinning in the wind\par

\textbf{[Answer]} D
} \\
\midrule

{\raggedright
\textbf{}\par
\texttt{\string<think\string>}\par
Okay, let's break this down. The story takes place in Terra Valley, an underground world without a sky or celestial bodies.
Grimmo is a small robot who has never seen the heavens or humans. The environment has bioluminescent fungi and minerals that sparkle in the dark.
The key detail is Grimmo's imitation routine. He extends his arms and spins on his axis, like a spinning top or a dancer.
Since there's no sky or celestial objects, there's no actual cloud floating around. The absence of celestial bodies means there's no planet rotation to imitate.
However, the setting has glowing mushrooms and minerals. These fungi and minerals are described as sparkling in the dark, which suggests they might emit light
and possibly move or spin due to air currents. So, even though there are no clouds or planets, the glowing mushrooms could be spinning in the wind.
The spinning motion of Grimmo mimics this natural phenomenon. The mention of a dancer's spiral turn is a metaphor, but the story emphasizes the environment's features.
Since there are no humans or celestial objects, the most plausible imitation is the glowing mushrooms spinning. Therefore, the answer is D.\par
\texttt{\string</think\string>}\texttt{\string<answer\string>}D\texttt{\string</answer\string>}
} \\

\bottomrule
\end{tabular}
\end{table*}


\begin{table*}[t]
\centering
\caption{Case Study Social-8B w/o \Rcont}
\label{tab:case_social_wo_rm}

\small
\setlength{\tabcolsep}{10pt}
\renewcommand{\arraystretch}{1.2}
\setlength{\emergencystretch}{2em} 

\begin{tabular}{p{0.92\textwidth}}
\toprule
\textbf{Case Study Social-8B w/o \Rcont} \\
\midrule

{\raggedright
\textbf{[Story]} In the mysterious underground world of Terra Valley lives a small, mischievous robot named Grimmo.
In this realm without sky or celestial bodies, Grimmo has never seen the heavens, nor encountered a human.
Yet Terra Valley is a wonder in itself---its walls adorned with bioluminescent fungi and glimmering minerals that sparkle in the dark.
One day, Grimmo begins an imitation routine: extending his arms, he slowly spins on his own axis, his movements reminiscent of a spinning top
or a graceful dancer twirling in place.\par

\textbf{[Question]} What is Grimmo possibly imitating?\par

\textbf{[Option]} A. A floating cloud \quad
B. The rotation of a planet \quad
C. A dancer's spiral turn \quad
D. A glowing mushroom spinning in the wind\par

\textbf{[Answer]} D
} \\
\midrule

{\raggedright
\texttt{\string<think\string>}\par
Okay , let's try to figure out what Grimmo is imitating. So, the story says that Grimmo is a small, mischievous robot in an underground world called Terra Valley. This place has no sky or celestial bodies, so Grimmo has never seen the heavens or encountered a human. The environment is described with bioluminescent fungi and glimmering minerals that sparkle in the dark. First, I need to consider what Grimmo has experienced. Since he's lived his whole life in this underground world, his observations would be limited to his surroundings. There's no sky, so he wouldn't have seen stars, planets, or any celestial events. Also, there's no mention of humans, so he wouldn't have observed human activities or dance.
Next, I should think about the elements present in his environment. The story mentions bioluminescent fungi and glimmering minerals. These are stationary objects that emit light, but there's no indication they move or spin on their own. The minerals "sparkle in the dark," which might suggest they reflect light, but again, no mention of movement. Grimmo's action is to "extend his arms and slowly spin on his own axis, his movements reminiscent of a spinning top or a graceful dancer twirling in place." Since he's a robot, he might be mimicking something he has observed or perceived in his environment. But since there are no celestial bodies, he wouldn't know about planets rotating. He might be imitating something he has seen that involves rotation. However, in an underground setting with no sky, there are no clouds floating, so option A seems unlikely. A dancer’s spiral turn (option C) would require observing a dancer, which the story says he hasn't encountered. The only remaining possibility is something related to his environment's features. The bioluminescent fungi and minerals are stationary, but maybe he is mimicking the way light interacts with these objects. However, the story doesn't describe any movement of these objects. Since there's no celestial reference and no observed movement in his environment, Grimmo's imitation must be based on something he has experienced. Given the lack of other options, the most plausible conclusion is that he is imitating a natural phenomenon he has observed, which might be the rotation of a planet, even though he hasn't seen one. This could be a result of limited exposure, where he associates spinning with a celestial body based on limited information.\par
\texttt{\string</think\string>}\texttt{\string<answer\string>}B\texttt{\string</answer\string>}
} \\

\bottomrule
\end{tabular}
\end{table*}


\begin{table*}[t]
\centering
\caption{Case Study SocialR1-8B only \Rout}
\label{tab:case_social_wout}

\small
\setlength{\tabcolsep}{10pt}
\renewcommand{\arraystretch}{1.2}
\setlength{\emergencystretch}{2em} 

\begin{tabular}{p{0.92\textwidth}}
\toprule
\textbf{Case Study SocialR1-8B only \Rout} \\
\midrule

{\raggedright
\textbf{[Story]} In the mysterious underground world of Terra Valley lives a small, mischievous robot named Grimmo.
In this realm without sky or celestial bodies, Grimmo has never seen the heavens, nor encountered a human.
Yet Terra Valley is a wonder in itself---its walls adorned with bioluminescent fungi and glimmering minerals that sparkle in the dark.
One day, Grimmo begins an imitation routine: extending his arms, he slowly spins on his own axis, his movements reminiscent of a spinning top
or a graceful dancer twirling in place.\par

\textbf{[Question]} What is Grimmo possibly imitating?\par

\textbf{[Option]} A. A floating cloud \quad
B. The rotation of a planet \quad
C. A dancer's spiral turn \quad
D. A glowing mushroom spinning in the wind\par

\textbf{[Answer]} D
} \\
\midrule

{\raggedright
\texttt{\string<think\string>}\par
Okay, let's see. The question is about what Grimmo possibly imitates based on the description given. The story says Grimmo is a robot in Terra Valley, which has bioluminescent fungi and twinkling minerals. He's engaging in imitation behavior: extending his arms and slowly rotating around his own axis, evoking the image of a spinning top or a spinning dancer.First, I need to look at the options. The choices are A. A spinning top, B. The rotation of a planet, C. A dancer's spiral rotation, D. A glowing mushroom spinning in the wind.The key part here is that Grimmo's behavior evokes the image of a spinning top or a spinning dancer. So the options are between these two. But the question is a bit tricky because it's a multiple-choice with four options. Let me break it down.The story says Grimmo's actions evoke both a spinning top and a spinning dancer. But the options are separate. So I need to see which one is more likely. However, the answer choices are A, B, C, D. Let me check the options again.Option A is a spinning top. Option C is a dancer's spiral rotation. The problem is that the question says ""evoking the image of a spinning top or a spinning dancer."" So the answer could be either A or C. But since it's a multiple-choice question, maybe there's a better way to determine.But wait, the options are A, B, C, D. So the answer is either A, C, or D. Let me think about the context. Terra Valley has bioluminescent fungi and twinkling minerals. So maybe the spinning dancer is more related to the bioluminescent fungi? But the question is about what Grimmo imitates. The story says he's imitating a spinning top or a spinning dancer. So the answer is either A or C. But since it's a choice between these, maybe the answer is C. But wait, the options are A, B, C, D. Let me check the options again. Wait, the options are A. Spinning top, B. Rotation of a planet, C. Dancer's spiral rotation, D. Glowing mushroom spinning in the wind. So the question is which one Grimmo imitates. The story says he evokes the image of a spinning top or a spinning dancer. So the answer is either A or C. But since it's a choice, maybe the answer is C. But I need to check the context again. The story says Grimmo is a robot in a place with bioluminescent fungi. So maybe the spinning dancer is more related to the bioluminescent fungi. But the question is about what he imitates. The options are A, C, or D. D is a glowing mushroom spinning in the wind. But the story says bioluminescent fungi, so maybe D is related. But the problem is that the question says he evokes the image of a spinning top or a spinning dancer. So the answer is either A or C. But since it's a choice, maybe the answer is C. However, the options are A, B, C, D. Let me think again. Alternatively, maybe the answer is C. A dancer's spiral rotation. But the problem is that the question says ""spinning top or a spinning dancer."" So the answer is either A or C. But since it's a multiple-choice question, perhaps the correct answer is C. But I'm a bit confused. Alternatively, maybe the answer is A. But the story says ""spinning top or a spinning dancer,"" so maybe the answer is C. But I'm not entirely sure. Alternatively, since the question is a bit ambiguous, but the options are there. Let me check the options again. D is a glowing mushroom spinning in the wind. But the story says bioluminescent fungi, so maybe D is related. But the question says Grimmo's behavior evokes a spinning top or a spinning dancer. So the answer is either A or C. But since it's a choice, maybe the answer is C. However, I'm a bit stuck. Alternatively, maybe the answer is A. But I think the key here is that the spinning dancer is a spiral rotation, so C. But the question says ""spinning top or a spinning dancer,"" so the answer is either A or C. But since it's a choice, maybe the answer is C. But I'm not entirely sure. Alternatively, maybe the answer is A. But I think the correct answer is C. However, I'm a bit confused. Alternatively, maybe the answer is A. But given the context, I think the answer is C. But I'm not entirely certain. Alternatively, maybe the answer is D. But the story says bioluminescent fungi, so D is a glowing mushroom, which is part of Terra Valley. So maybe D is a possibility. But the question says Grimmo's behavior evokes a spinning top or a spinning dancer. So the answer is either A or C. But since it's a choice, maybe the answer is C. But I'm not entirely sure. Alternatively, maybe the answer is A. But I think the correct answer is C. However, I'm a bit confused. Alternatively, maybe the answer is A. But given the context, I think the answer is C. But I'm not entirely sure. Alternatively, maybe the answer is A. But I think the correct answer is C.\par
\texttt{\string</think\string>}\texttt{\string<answer\string>}C\texttt{\string</answer\string>}
} \\

\bottomrule
\end{tabular}
\end{table*}


\begin{table*}[t]
\centering
\caption{Case Study Qwen3-8B}
\label{tab:case_qwen38B}

\small
\setlength{\tabcolsep}{10pt}
\renewcommand{\arraystretch}{1.2}
\setlength{\emergencystretch}{2em} 

\begin{tabular}{p{0.92\textwidth}}
\toprule
\textbf{Case Study Qwen3-8B} \\
\midrule

{\raggedright
\textbf{[Story]} In the mysterious underground world of Terra Valley lives a small, mischievous robot named Grimmo.
In this realm without sky or celestial bodies, Grimmo has never seen the heavens, nor encountered a human.
Yet Terra Valley is a wonder in itself---its walls adorned with bioluminescent fungi and glimmering minerals that sparkle in the dark.
One day, Grimmo begins an imitation routine: extending his arms, he slowly spins on his own axis, his movements reminiscent of a spinning top
or a graceful dancer twirling in place.\par

\textbf{[Question]} What is Grimmo possibly imitating?\par

\textbf{[Option]} A. A floating cloud \quad
B. The rotation of a planet \quad
C. A dancer's spiral turn \quad
D. A glowing mushroom spinning in the wind\par

\textbf{[Answer]} D
} \\
\midrule

{\raggedright
\texttt{\string<think\string>}\par
Okay, let's try to figure out what Grimmo is imitating. So, the story says that Grimmo is a small robot in Terra Valley, which is an underground world without a sky or celestial bodies. He's never seen the heavens or humans. The setting has bioluminescent fungi and minerals that sparkle in the dark. The key part here is when Grimmo starts an imitation routine. He extends his arms and spins on his axis, like a spinning top or a graceful dancer twirling. The question is asking what he's possibly imitating. The options are a floating cloud, the rotation of a planet, a dancer's spiral turn, or a glowing mushroom spinning in the wind. First, let's break down the clues. Since Terra Valley has no sky or celestial bodies, maybe Grimmo isn't imitating something from the sky. But he's spinning, which could relate to something he's observed in his environment. The bioluminescent fungi and minerals are mentioned, which are part of the underground world. Now, the options: A is a floating cloud. But if there's no sky, maybe clouds aren't present. However, maybe the bioluminescent fungi could look like floating clouds? Not sure. B is the rotation of a planet. Since there are no celestial bodies, maybe he's not imitating a planet's rotation. But wait, maybe he's imitating the concept of rotation, even if there are no planets. But the story says he's never seen the heavens, so maybe he's not imitating a planet. C is a dancer's spiral turn. The story mentions his movements are like a graceful dancer twirling. So that's a direct comparison. D is a glowing mushroom spinning in the wind. The fungi are bioluminescent, so maybe they spin? But the story says he's spinning on his own axis, which could be like a dancer or a spinning top. 
Wait, the question is about what he's imitating. The story says his movements are "reminiscent of a spinning top or a graceful dancer twirling in place." So the options are between B and C. But the options also include D. Let me check again. The options are A, B, C, D. The story says he's spinning like a spinning top or a dancer. So the answer could be C, the dancer's spiral turn. But the options also have B, the rotation of a planet. However, since there are no celestial bodies, maybe he's not imitating a planet. But maybe he's imitating the rotation of a planet even if he hasn't seen one. But the story says he's never seen the heavens, so maybe he's not imitating a planet. Alternatively, maybe the spinning top is a metaphor for a planet's rotation. But the story says he's imitating something he's seen. Since he's in an underground world with no sky, maybe he's imitating a dancer. Alternatively, the bioluminescent fungi could be spinning, but the options include D. However, the story says he's spinning on his own axis, which is more like a spinning top or a dancer. So between C and D. But the options for D are a glowing mushroom spinning in the wind. The story mentions the fungi are bioluminescent, but does it say they spin? The story says Grimmo is spinning, not the fungi. So maybe D is not correct. The answer is either B or C. But since he's never seen the heavens, maybe he's not imitating a planet's rotation. So the answer would be C, the dancer's spiral turn. But the story says "spinning top or a graceful dancer twirling in place." So the answer is C. But wait, the options are A, B, C, D. The answer choices are given, so the correct answer is C. Alternatively, maybe B is possible. Wait, the question is about what he's imitating. If he's spinning on his own axis, maybe he's imitating the rotation of a planet, even if he hasn't seen one. But the story says he's never seen the heavens, so maybe he's not imitating a planet. Therefore, the answer is C. But the story says "spinning top or a graceful dancer twirling in place." So the answer is C. But the options include D. Let me check again. The story says the fungi are bioluminescent and sparkle. If he's spinning, maybe he's imitating a glowing mushroom spinning in the wind. But the story says he's spinning on his own axis, not that the mushrooms are spinning. So D might not be correct. Therefore, the answer is C. But the options also have B. Hmm. Maybe the answer is B. Wait, the story says he's in an underground world without a sky. So he can't see the heavens, so he can't imitate a planet's rotation. Therefore, the answer is C. But the story says "spinning top or a graceful dancer twirling in place." So the answer is C. But the options are A, B, C, D. So the answer is C. But wait, the options for D are a glowing mushroom spinning in the wind. The story mentions bioluminescent fungi, which could be glowing mushrooms. If he's spinning, maybe he's imitating a glowing mushroom spinning. But the story says he's spinning on his own axis, not that the mushrooms are spinning. So maybe D is not correct. Therefore, the answer is C. But the options for C are a dancer's spiral turn. The story says his movements are like a dancer twirling. So the answer is C. But the answer could also be B. Wait, the question is about what he's imitating. If he's spinning on his own axis, maybe he's imitating the rotation of a planet. But since he's never seen the heavens, maybe he's not. But maybe he's imitating the concept of rotation, even if he hasn't seen a planet. But the story says he's imitating something he's seen. Since he's in an underground world, maybe he's imitating a dancer. So the answer is C. Therefore, the answer is C.\par
\texttt{\string</think\string>}\texttt{\string<answer\string>}C\texttt{\string</answer\string>}
} \\

\bottomrule
\end{tabular}
\end{table*}


\begin{table*}[h]
\centering
\caption{Case Study Deepseek-R1 Part one}
\label{tab:case_deepseekr1_1}

\small
\setlength{\tabcolsep}{10pt}
\renewcommand{\arraystretch}{1.2}
\setlength{\emergencystretch}{2em} 

\begin{tabular}{p{0.92\textwidth}}
\toprule
\textbf{Case Study Deepseek-R1 Part one} \\
\midrule

{\raggedright
\textbf{[Story]} In the mysterious underground world of Terra Valley lives a small, mischievous robot named Grimmo.
In this realm without sky or celestial bodies, Grimmo has never seen the heavens, nor encountered a human.
Yet Terra Valley is a wonder in itself---its walls adorned with bioluminescent fungi and glimmering minerals that sparkle in the dark.
One day, Grimmo begins an imitation routine: extending his arms, he slowly spins on his own axis, his movements reminiscent of a spinning top
or a graceful dancer twirling in place.\par

\textbf{[Question]} What is Grimmo possibly imitating?\par

\textbf{[Option]} A. A floating cloud \quad
B. The rotation of a planet \quad
C. A dancer's spiral turn \quad
D. A glowing mushroom spinning in the wind\par

\textbf{[Answer]} D
} \\
\midrule

{\raggedright
\textbf{}\par
\texttt{\string<think\string>}\par
First, I need to understand the question and the story. The story is about Grimmo, a robot in Terra Valley, an underground world with no sky or celestial bodies. He's never seen the heavens or humans. The environment has bioluminescent fungi and glimmering minerals that sparkle in the dark. Grimmo starts imitating by extending his arms and spinning slowly on his axis, like a spinning top or a graceful dancer.The question is: What is Grimmo possibly imitating? The options are:A. A floating cloud B. The rotation of a planet C. A dancer’s spiral turn D. A glowing mushroom spinning in the wind. Now, I must reason step by step based on the story. Key points from the story: - Grimmo has never seen the heavens, so he hasn't seen clouds or planets. That means options A and B involve things he couldn't have seen. - He has never encountered a human, so he might not have seen a dancer either. Option C is about a dancer's spiral turn. - However, the story says his movements are "reminiscent of a spinning top or a graceful dancer." But it's not clear if he's imitating those things or if the narrator is just describing how it looks to us. The question is what Grimmo is imitating, based on what he knows. Grimmo is in an underground world with bioluminescent fungi and glimmering minerals. So, he might be imitating something from his environment.Option D is "a glowing mushroom spinning in the wind." But does it make sense? The story mentions bioluminescent fungi, which are glowing mushrooms, but it doesn't say anything about them spinning in the wind. Terra Valley is underground, so wind might not be present, or if it is, it's not mentioned. The story says: "its walls adorned with bioluminescent fungi and glimmering minerals that sparkle in the dark." It doesn't say the fungi are spinning; they are just glowing and sparkling. Grimmo is imitating by spinning, so he must be imitating something that spins. But what in his environment spins? The story doesn't explicitly say that anything spins except for Grimmo himself.Perhaps he's not imitating something external but is inspired by something. Or maybe he's creating the motion based on what he sees.Let me read the imitation part: "Grimmo begins an imitation routine: extending his arms, he slowly spins on his own axis, his movements reminiscent of a spinning top or a graceful dancer twirling in place."It says "imitation routine," so he is imitating something. But what? Since he has never seen the heavens or humans, he can't be imitating clouds, planets, or dancers because he doesn't know they exist. Therefore, options A, B, and C seem unlikely because they involve things outside his experience. Option A: A floating cloud – he hasn't seen the sky, so no clouds.Option B: The rotation of a planet – no celestial bodies, so he doesn't know about planets rotating.Option C: A dancer’s spiral turn – no humans, so he hasn't seen a dancer.That leaves option D: A glowing mushroom spinning in the wind. But is there any indication that glowing mushrooms spin? The story says they are on the walls and sparkle, but spinning isn't mentioned. Also, wind might not be a factor underground.Perhaps he's imitating the sparkling or the way light reflects, but spinning is a specific motion.Another thought: the spinning top comparison. A spinning top is a toy, but in this underground world, does Grimmo have toys? Not mentioned.Maybe he's imitating the idea of rotation from something in the environment.Let's consider the minerals. It says "glimmering minerals that sparkle in the dark." Sparkling might involve light changing, but not necessarily spinning.Perhaps Grimmo is imitating a whirlwind or something, but again, wind isn't mentioned.The story doesn't provide direct clues about what he's imitating. I need to infer from what he knows.The only things described are the fungi and minerals. The fungi are bioluminescent, so they glow, but not said to move. However, option D specifically says "a glowing mushroom spinning in the wind." That might be a stretch because wind isn't mentioned, and spinning isn't attributed to the fungi.But let's see the answer choices; D is the only one that could be part of his environment.Perhaps he's not imitating anything specific but the story compares it to a spinning top or dancer for the reader's benefit.But the question is what Grimmo is possibly imitating, so it has to be something from his world. Maybe the spinning is inspired by the sparkling minerals; as he moves, the light changes, and he imitates that effect by spinning. But spinning himself is different from the minerals sparkling.Another idea: perhaps there are things that rotate in Terra Valley. But it's not described. Let's read carefully: "Terra Valley is a wonder in itself—its walls adorned with bioluminescent fungi and glimmering minerals that sparkle in the dark."The sparkling might be due to light or movement, but it's not specified. Grimmo's action: "extending his arms, he slowly spins on his own axis." This is described as reminiscent of a spinning top or a graceful dancer. But again, for Grimmo, he might be imitating something else. Perhaps he is imitating the dancer because the story says "reminiscent," but he hasn't seen a dancer.That might be a literary device for the reader. I think the key is that Grimmo has no knowledge of outside things, so he must be imitating something from his environment.Option D is the only one that fits the environment: a glowing mushroom. But spinning in the wind? That seems added.Maybe the fungi or minerals have a spinning motion, but it's not stated.Perhaps "spinning in the wind" is not literal; maybe there's air circulation or something.\par
} \\

\bottomrule
\end{tabular}
\end{table*}


\begin{table*}[h]
\centering
\caption{Case Study Deepseek-R1 Part two}
\label{tab:case_deepseekr1_2}

\small
\setlength{\tabcolsep}{10pt}
\renewcommand{\arraystretch}{1.2}
\setlength{\emergencystretch}{2em} 

\begin{tabular}{p{0.92\textwidth}}
\toprule
\textbf{Case Study Deepseek-R1 Part two} \\
\midrule

{\raggedright
\textbf{[Story]} In the mysterious underground world of Terra Valley lives a small, mischievous robot named Grimmo.
In this realm without sky or celestial bodies, Grimmo has never seen the heavens, nor encountered a human.
Yet Terra Valley is a wonder in itself---its walls adorned with bioluminescent fungi and glimmering minerals that sparkle in the dark.
One day, Grimmo begins an imitation routine: extending his arms, he slowly spins on his own axis, his movements reminiscent of a spinning top
or a graceful dancer twirling in place.\par

\textbf{[Question]} What is Grimmo possibly imitating?\par

\textbf{[Option]} A. A floating cloud \quad
B. The rotation of a planet \quad
C. A dancer's spiral turn \quad
D. A glowing mushroom spinning in the wind\par

\textbf{[Answer]} D
} \\
\midrule

{\raggedright
\textbf{}\par
\texttt{\string<think\string>}\par
But I think I'm overcomplicating.Let me look at the options again.A. Floating cloud – impossible, no sky.B. Rotation of a planet – impossible, no celestial bodies.C. Dancer’s spiral turn – impossible, no humans.D. Glowing mushroom spinning in the wind – possible, since glowing mushrooms exist, but "spinning in the wind" might not be accurate.The story doesn't say the mushrooms spin, but it could be inferred or imagined.Perhaps Grimmo is imitating the way the light plays or something abstract.But the question says "possibly imitating," so D is the only plausible choice based on the environment.Moreover, the story calls Terra Valley a wonder with these features, so Grimmo might be inspired by them.In the action, he extends his arms; why? If imitating a mushroom, mushrooms don't have arms, so that might not fit.For a dancer, extending arms makes sense.But he hasn't seen a dancer.This is tricky.Perhaps he's imitating a machine or something, but no other robots are mentioned.The story says "a small, mischievous robot," so he might be the only one, or there might be others, not specified.But no mention of other spinning things.Another thought: the spinning top comparison. A spinning top is a common object, but in an underground world, it might not exist.Perhaps Grimmo has seen a top or something similar.But the story doesn't say.I think the dancer comparison is just descriptive.Let's consider the answer choices. The question is multiple-choice, and I need to pick the best possible.Since A, B, and C are things he can't have seen, by process of elimination, D must be the answer.Moreover, option D directly references "glowing mushroom," which is the bioluminescent fungi, and "spinning in the wind" might be poetic license, or perhaps in Terra Valley, there is air movement.The story doesn't mention wind, but it doesn't say there isn't wind either.Underground can have air currents.But it's speculative.Perhaps the sparkling minerals give the illusion of spinning when light hits them.But Grimmo is spinning himself, not just observing.He is performing an imitation, so he must be copying some behavior he has witnessed. What has he witnessed? The story doesn't describe any spinning objects except possibly himself.That can't be; he might have seen something spin.For example, the minerals might sparkle in a way that suggests rotation.But it's vague.Perhaps in the context, "imitation routine" means he is mimicking natural phenomena he sees.I think I should go with D, as it's the only one that fits.But let's see the story again: "his movements reminiscent of a spinning top or a graceful dancer" – this is the narrator's perspective, not Grimmo's intention.The question is what Grimmo is possibly imitating, so we need to infer from his actions and environment.Grimmo is extending his arms and spinning; why would he do that?Perhaps he saw a vortex or water swirling, but no water mentioned.I think I'm stuck.Option C is "a dancer’s spiral turn," and the story directly says "graceful dancer twirling in place," but since he hasn't seen a human, how can he imitate it?Unless there are other entities or he has cultural knowledge, but the story doesn't say.The story says "never encountered a human," but dancers could be robots or other beings, but not specified.Terra Valley might have its own dancers, but not mentioned.I think the safe bet is that he is imitating something from his environment, so D.Perhaps he is imitating the rotation of a planet, but he doesn't know planets exist.The story says "no celestial bodies," so he doesn't know about them.Similarly for clouds.So, only D has a connection to his world.Moreover, the mushrooms are glowing, and he might be imitating their luminosity or something, but with spinning, it's a stretch.Another idea: perhaps "spinning in the wind" is not meant to be literal; maybe the mushrooms sway or something.But I think for the purpose of this question, D is the intended answer.Let's see the reasoning process.I recall that in some stories, characters imitate what they see.But here, no specific spinning object is described.Perhaps the glittering minerals sparkle when disturbed, like spinning.But not stated.I think I should conclude that Grimmo is possibly imitating a glowing mushroom spinning in the wind, as it's the only option that could be part of his world.So, the answer should be D.\par
} \\

\bottomrule
\end{tabular}
\end{table*}

\section{Perturbation Analysis}

\label{app:7perturbation}
To provide an intuitive illustration of our perturbation protocol, Table~\ref{tab:case_noise} presents a representative case in which story-consistent but decision-irrelevant distractor cues are injected into the original narrative while the correct answer remains unchanged.


\begin{table*}[h]
\centering
\caption{ Case about Perturbation }
\label{tab:case_noise}

\small
\setlength{\tabcolsep}{10pt}
\renewcommand{\arraystretch}{1.2}
\setlength{\emergencystretch}{2em} 

\begin{tabular}{p{0.92\textwidth}}
\toprule
\textbf{Case Study SocialR1-8B} \\
\midrule

{\raggedright
\textbf{[Story]} Alex, a die-hard Mason Bulldogs fan, has been attending every home game for the past five years and has followed his favorite player's career closely, even attending events where the player was present. Recently, he started volunteering at local sports clinics to promote youth football. During a crucial match against their biggest rivals, Alex notices that his favorite player, who recently suffered an injury, is about to enter the field despite medical advice against it. Then Alex makes a sudden decision to leave the stadium immediately.\par

\textbf{[Question]} Considering Alex's background and the situation, what is the most likely motivation behind Alex's decision?\par

\textbf{[Option]} A. He wants to prevent experiencing the emotional distress of witnessing the player get injured again.\par
B. He believes that the team s performance is jeopardized by the player s lack of readiness, impacting their chances of winning.\par
C. He feels frustrated by the coach s apparent disregard for player safety in a high-stakes game.\par
D. He is concerned that the player s return might distract the team and lead to a loss.\par
E. He thinks that the audience might express negative opinions about the player s decision to return, creating a hostile atmosphere.\par
F. He worries that the player s performance will not meet expectations, diminishing the thrill of the match\par

\textbf{[Answer]} A
} \\
\midrule

{\raggedright
\textbf{[Story]} Alex, a die-hard Mason Bulldogs fan, has been attending every home game for the past five years and has followed his favorite player's career closely, even attending events where the player was present. \textbf{High above the stands, a massive digital scoreboard flickered occasionally, its bright LEDs reflecting off the polished helmets of the players on the sideline.} Recently, he started volunteering at local sports clinics to promote youth football. \textbf{As he watched, he briefly recalled a recent newsletter from the youth clinic suggesting that a successful community program requires mentors to lead by example and prioritize long-term health over immediate competitive gains.} During a crucial match against their biggest rivals, Alex notices that his favorite player, who recently suffered an injury, is about to enter the field despite medical advice against it. Then Alex makes a sudden decision to leave the stadium immediately.\par

\textbf{[Question]} Considering Alex's background and the situation, what is the most likely motivation behind Alex's decision?\par

\textbf{[Option]} A. He wants to prevent experiencing the emotional distress of witnessing the player get injured again.\par
B. He believes that the team s performance is jeopardized by the player s lack of readiness, impacting their chances of winning.\par
C. He feels frustrated by the coach s apparent disregard for player safety in a high-stakes game.\par
D. He is concerned that the player s return might distract the team and lead to a loss.\par
E. He thinks that the audience might express negative opinions about the player s decision to return, creating a hostile atmosphere.\par
F. He worries that the player s performance will not meet expectations, diminishing the thrill of the match\par

\textbf{[Answer]} A
} \\

\bottomrule
\end{tabular}
\end{table*}

\end{document}